\documentclass[preprint,12pt]{elsarticle}

\usepackage{amssymb}
\usepackage{booktabs}  
\usepackage{amsmath}   
\usepackage{graphicx}  
\usepackage{subcaption} 
\usepackage{tcolorbox}
\usepackage{algorithm}
\usepackage{algorithmic}
\usepackage{comment}
\usepackage{soul}  
\usepackage{xcolor} 
\sethlcolor{yellow}
\usepackage{float}

\journal{Mechanism and Machine Theory}

\begin{document}

\begin{frontmatter}



\title{Backstepping Control of Tendon-Driven Continuum Robots in Large Deflections Using the Cosserat Rod Model}

 \author[label1]{Rana Danesh}
\author[label1]{Farrokh Janabi-Sharifi}
 \affiliation[label1]{organization={Mechanical, Industrial, and Mechatronics Engineering, Toronto Metropolitan University},
             addressline={350 Victoria St},
             city={Toronto},
             postcode={M5B 2K3},
             state={ON},
            country={Canada}}


\begin{abstract}
This paper presents a study on the backstepping control of tendon-driven continuum robots for large deflections using the Cosserat rod model. Continuum robots are known for their flexibility and adaptability, making them suitable for various applications. However, modeling and controlling them pose challenges due to their nonlinear dynamics. To model their dynamics, the Cosserat rod method is employed to account for significant deflections, and a numerical solution method is developed to solve the resulting partial differential equations. Previous studies on controlling tendon-driven continuum robots using Cosserat rod theory focused on sliding mode control and were not tested for large deflections, lacking experimental validation. In this paper, backstepping control is proposed as an alternative to sliding mode control for achieving a significant bending.
The numerical results are validated through experiments in this study, demonstrating that the proposed backstepping control approach is a promising solution for achieving large deflections with smoother trajectories, reduced settling time, and lower overshoot. Furthermore, two scenarios involving external forces and disturbances were introduced to further highlight the robustness of the backstepping control approach.

\end{abstract}


\begin{highlights}
\item Developed backstepping control for large deflections in tendon-driven robots.
\item Used Cosserat rod model for accurate dynamic modeling under significant deflections.
\item Numerical simulations show smooth trajectories using backstepping control.
\item Experimental results confirm superior performance over sliding mode control.

\end{highlights}


\end{frontmatter}



\section{Introduction}
\label{sec1}
In recent years, continuum robots (CRs) have gained significant popularity due to their inherent compliance and flexibility, which enable them to operate within unstructured environments \cite{trivedi2008}. Continuum robots offer numerous applications ranging from space exploration, search and rescue operations, and minimally invasive surgeries to aerial manipulation due to their unique properties \cite{jalali2024}. Despite CRs' advantages, their nonlinear dynamic model and flexible nature present significant challenges for modeling and control design. 

There are two frameworks for modeling CRs, which can be categorized into kinematics-based and dynamics-based approaches. Kinematic modeling provides straightforward models suitable for scenarios with minimal dynamic behavior. These methods are renowned and extensively utilized for modeling and controlling CRs. However, in more complex situations involving external forces or large deflection bending, dynamic modeling becomes essential for achieving an accurate representation \cite{burgner-kahrs2015}.

Cosserat rod theory is one of the dynamic methods that has gained popularity as a useful tool for creating general models of CRs. It does not rely on small-deflection approximations and can handle any nonlinear stress-strain relationship, making it broadly applicable to a wide range of CRs \cite{rucker2011}-\cite{dupont2009}.

One type of CRs that gain attention due to their simple structure and easy implementation are the tendon-driven continuum robots (TDCRs) \cite{walker2013}. Cosserat rod theory is utilized to model these types of CRs in numerous papers \cite{janabi-sharifi2021}-\cite{prisco1997}. 


A few studies have focused on controlling TDCRs using the Cosserat rod model. Alqumsan proposed a robust controller based on sliding mode for TDCRs, which are modeled using the Cosserat rod theory. They included control input saturation in the design process to prevent sudden changes in tendon tension \cite{alqumsan2019a}.
In their next paper, they designed a robust controller utilizing multi-surface sliding mode control to handle mismatched uncertainties in a TDCR \cite{alqumsan2019b}. Later, they combined the multi-surface sliding mode control technique with RBF neural networks as uncertainty approximators to address the effect of uncertainties on the robot's states and ensure tracking stability. In their papers, the simulation results demonstrated the effectiveness of the proposed control schemes for a $15\ $cm deflection of a TDCR with a length of $30\ $cm \cite{alqumsan2021}.

Samadikhoshkho et al. also designed sliding mode controller for their aerial manipulator involving TDCR. To model the robot, they employed Cosserat rod theory. The simulation results indicated that the proposed controller effectively tracks a trajectory with a $12\ $cm deflection for a TDCR with a length of $30\ $cm \cite{samadikhoshkho2020}.

Tang et al. conducted a study on the impact of actuator faults on TDCRs. They proposed an adaptive sliding mode fault-tolerant control scheme and modeled the robot using the Cosserat rod theory. Their defined trajectory for the $30\ $cm length robot had a deflection of $10\ $cm. The suggested control method demonstrated good performance in terms of trajectory tracking and ensured asymptotic stability, even in the presence of actuator faults \cite{tang2023}.

Pekris et al. designed a feedforward control strategy for a tendon-driven continuum robot, employing the Cosserat rod model with implicit time discretization. Their results indicate accurate tip position tracking, suggesting that the control method is effective for precise and efficient control. However, they did not evaluate the controller's performance in scenarios with large deflections \cite{Pekris}.

Previous studies on controlling TDCRs using Cosserat rod models have not addressed large deflections, where nonlinearities are more prominent. Achieving large deflections is crucial for TDCRs due to their diverse applications that require a wide range of motion. Additionally, the controllers developed in these studies were only evaluated through simulation analysis and lacked experimental validation.

In this study, we address the challenge of achieving large deflections in TDCRs. Our approach involves implementing backstepping control to drive the robot’s tip to a $90$\textdegree$\ $ deflection, achieving significant bending under normal conditions, as well as in the presence of external forces and disturbances.
 This approach is motivated by observations of sudden changes when sliding mode control, as conducted by previous researchers, is used for large deflections. To demonstrate the effectiveness of backstepping control for TDCRs in achieving a large bending, an experimental setup is designed and implemented as part of this research.

To provide a guide to the organization of this paper, Section \ref{sec2} utilizes the Cosserat rod model to derive the dynamic equations of motion for a TDCR, while Section \ref{sec3} presents a numerical solution framework. Section \ref{sec4} introduces the two control methods used in this study, sliding mode and backstepping control.  Following this, in Section \ref{sec5}, the simulation results are reported, and the effectiveness of the approaches is evaluated with experiments. Finally, in Section \ref{sec6}, the study is concluded, and its remarkable findings are highlighted.

\section{Dynamic Model}
\label{sec2}

\begin{figure}[H]
  \centering
  \includegraphics[width=0.6\textwidth]{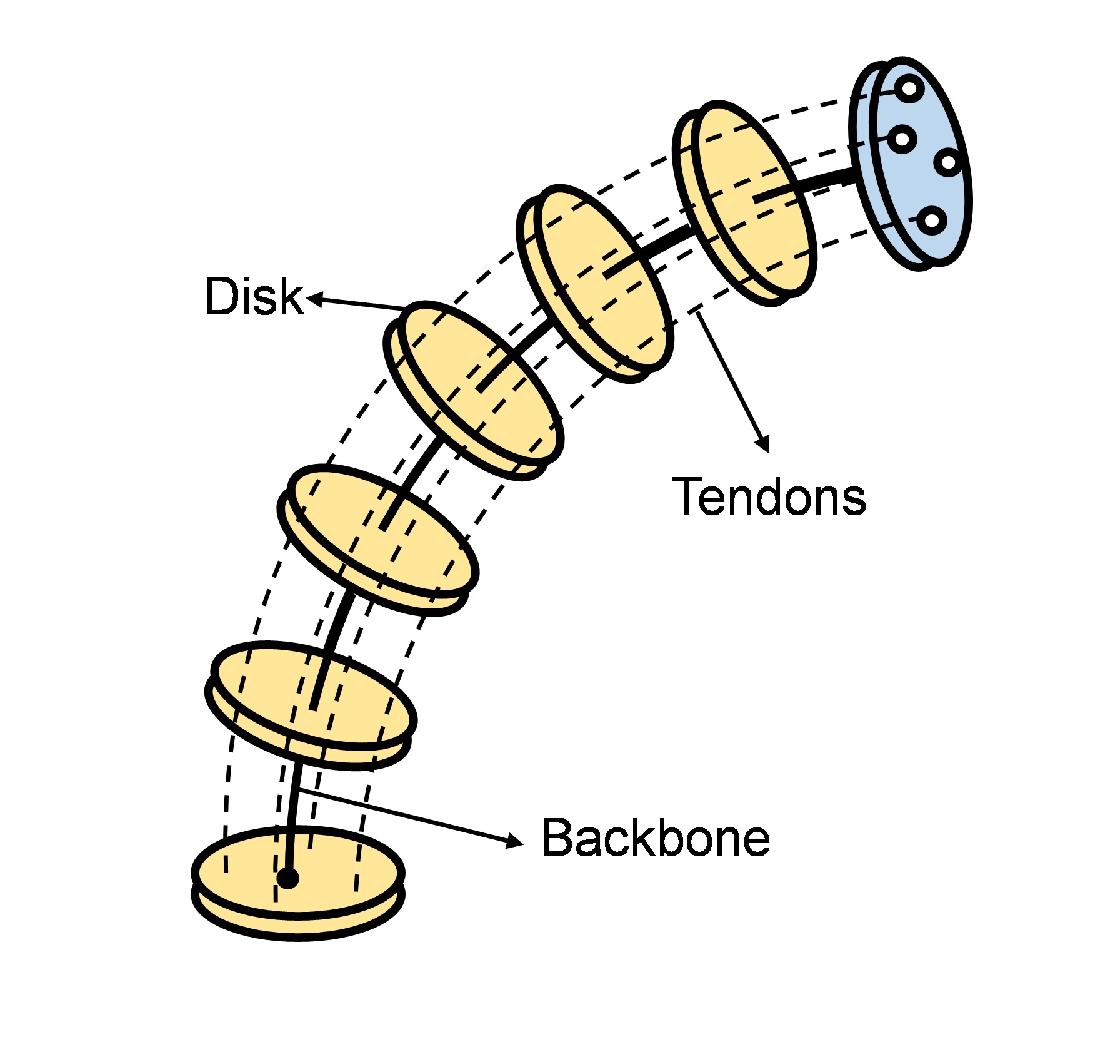}
  \caption{Schematic configuration of a single-section tendon-driven continuum robot.}
  \label{fig1}
\end{figure}

In a CR, there is an elastic backbone with separate disks that have holes for tendons to pass through. Tendon tensions cause forces and moments on the robot's backbone, which cause it to move. Fig.\ref{fig1} shows a schematic configuration of a TDCR.

\begin{figure}[H]
\centerline{\includegraphics[width=0.85\textwidth]{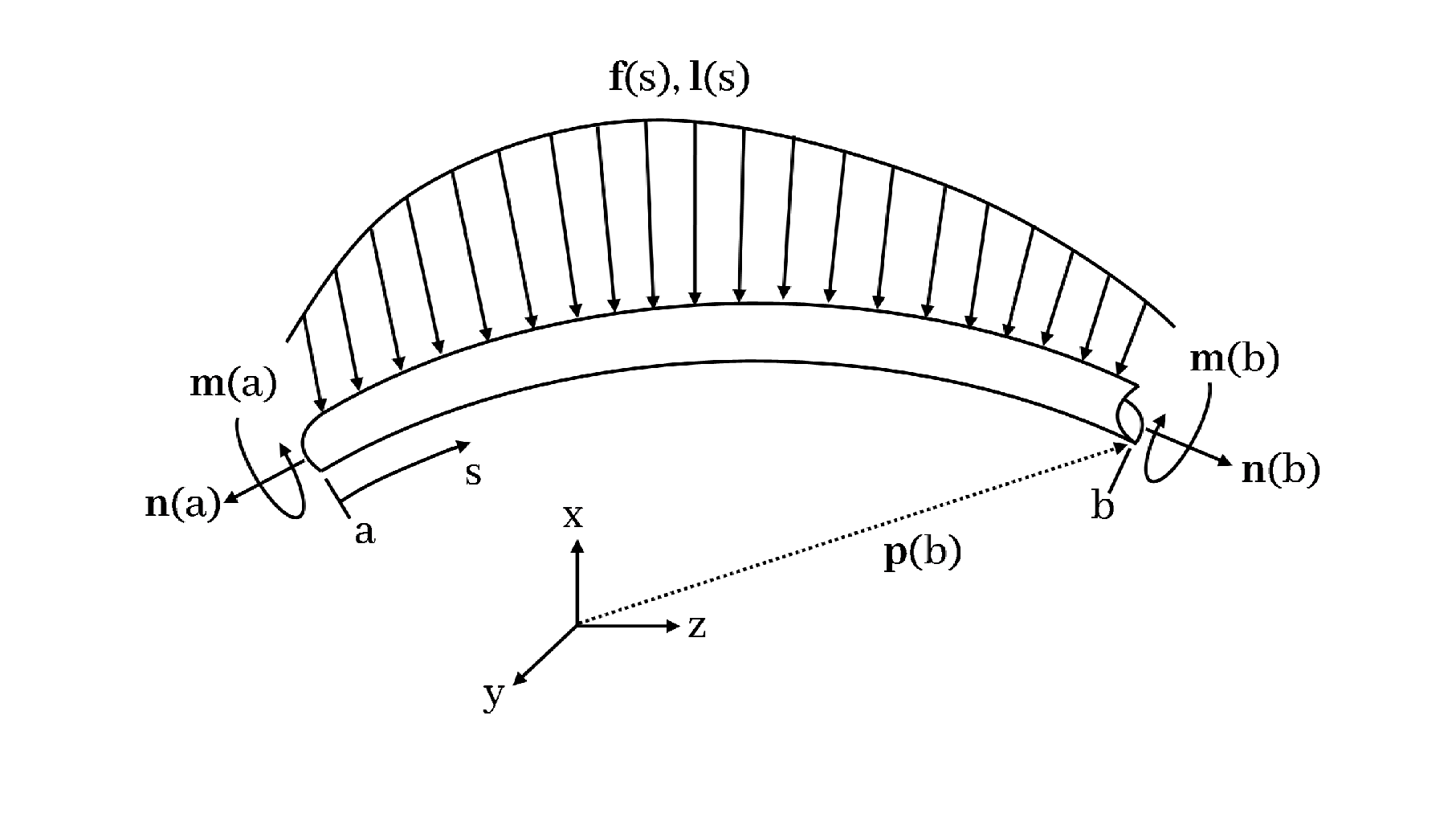}}
\caption{A section of a rod subjected to distributed forces $\mathbf{f}(s)$ and moments $\mathbf{l}(s)$ at a given time $t$.}
\label{fig2}
\end{figure}

As CRs have a high ratio of length to cross-section area, they can be modeled as rods. The Cosserat rod model is used to model the CR dynamics, and the state variables are determined by time $t\in \mathbb{R}$ and arc length $s\in \mathbb{R}$ \cite{rucker2011,janabi-sharifi2021,till2019}. Position and orientation of the rod can be described by a curve in space $\mathbf{p}(s,t)\in \mathbb{R}^3$ and a set of orthogonal directors ($\mathbf{d_1}$, $\mathbf{d_2}$, and $\mathbf{d_3}$) at each point $s$. Directors $\mathbf{d_1}$ and $\mathbf{d_2}$ span the cross section and $\mathbf{d_3}=\mathbf{d_1}\times \mathbf{d_2}$. We define the rotational matrix relating the local frame created by the directors to a global frame $\mathbf{R}(s,t) \in SO(3)$, the rate of change of position with respect to arc length $\mathbf{v}(s,t)$, the curvature $\mathbf{u}(s,t)$, linear velocity $\mathbf{q}(s,t)$, and angular velocity $\boldsymbol{\omega}(s,t)$. Also, the internal forces and moments are represented by $\mathbf{n}(s,t)\in \mathbb{R}^3$ and $\mathbf{m}(s,t)\in \mathbb{R}^3$, respectively. An arbitrary section of the rod with its corresponding parameters is depicted in Fig. \ref{fig2}.

To simplify the formulation of dynamic equations, the following two assumptions are commonly made \cite{janabi-sharifi2021}:

\begin{itemize}
    \item Every tendon’s tension is constant along its length; and
    \item Tendons remain at the same location in the cross section during deformation.
\end{itemize}

In global coordinates, TDCR partial differential equations (PDEs) are represented as follows:

\begin{equation}\label{PDE}
\left\{
\begin{aligned}
&\mathbf{p}_s=\mathbf{Rv},\ \mathbf{R}_s=\mathbf{R}\hat{\boldsymbol{u}},\\
&\mathbf{q}_s=\mathbf{v}_t \hat{\boldsymbol{u}}\mathbf{q}+\hat{\boldsymbol{\omega}}\mathbf{v},\ \boldsymbol{\omega}_s=\mathbf{u}_t-\hat{\boldsymbol{u}}\boldsymbol{\omega}\\
&\mathbf{n}_s=\rho A\boldsymbol{R}(\hat{\boldsymbol{\omega}}\mathbf{q}+\mathbf{q}_t)-\mathbf{f}\\
&\mathbf{m}_s=(\rho\boldsymbol{ RJ}\boldsymbol{\omega})_t -\hat{\mathbf{p}}_s\mathbf{n}-\mathbf{l}\\
&\mathbf{p}_t=\mathbf{Rq},\ \mathbf{R}_t=\mathbf{R}\hat{\boldsymbol{\omega}}
\end{aligned}
\right.
\end{equation}

\noindent Here $\hat{(.)}$ maps $\mathbb{R}^3$ to $SO(3)$, and as an example:

\begin{equation}\label{hat}
\mathbf{a}\in \mathbb{R}^3 \stackrel{\hat{(.)}}{\longrightarrow} \hat{\mathbf{a}}= 
\begin{bmatrix}
    0 & -a_3 & a_2 \\
    a_3 & 0 & -a_1 \\
    -a_2 & a_1 & 0
\end{bmatrix},(\hat{\mathbf{a}} \check{)}  = \mathbf{a}
\end{equation}

In (\ref{PDE}), $\rho$, $A$, and $\mathbf{J}$ represent density of backbone, cross section area, and rotational inertia matrix, respectively. $\mathbf{f}$ and $\mathbf{l}$ are general distributed forces and moments that can be defined as:

\begin{equation}\label{f_and_l}
\left\{
\begin{aligned}
&\mathbf{f}=\mathbf{f}_e+\mathbf{f}_{\text{tendon}}\\
&\mathbf{l}=\mathbf{l}_e+\mathbf{l}_{\text{tendon}}\\
\end{aligned}
\right.
\end{equation}

\noindent where $\mathbf{f}_e$ and $\mathbf{l}_e$ are external distributed force and moment, while $\mathbf{f}_{\text{tendon}}$ and $\mathbf{l}_{\text{tendon}}$  denote tendon tension’s distributed force and moment. We assume that $\mathbf{f}_e$ is the sum of gravitational and damping forces as $\rho A\mathbf{g}-
\mathbf{RCq\odot|q|}$, where $\mathbf{C}$ is the square law drag damping coefficient and $\odot$ is the Hadamard product. The force and moment due to tendon tensions can be represented as:

\begin{equation}\label{f_tendon}
\left\{
\begin{aligned}
&\mathbf{f}_{\text{tendon}}=\mathbf{R}(\mathbf{a}+\mathbf{A}\mathbf{v}_s+\mathbf{G}\mathbf{u}_s)\\
&\mathbf{l}_{\text{tendon}}=\mathbf{R}(\mathbf{b}+\mathbf{B}\mathbf{v}_s+\mathbf{H}\mathbf{u}_s)\\
\end{aligned}
\right.
\end{equation}

\noindent in which,

\begin{equation}\label{a_and_b}
\left\{
\begin{aligned}
&\mathbf{A}_i=-{T}_i\frac{(\hat{\mathbf{p}}^b_{is})^2}{||\mathbf{p}^b_{is}||^3},\mathbf{B}_i=\hat{\mathbf{r}}_i\mathbf{A}_i, \mathbf{A}=\sum_{i=1}^{n}\mathbf{A}_i, \mathbf{B}=\sum_{i=1}^{n}\mathbf{B}_i\\
&\mathbf{G}=-\sum_{i=1}^{n}\mathbf{A}_i \hat{\mathbf{r}}_i, \mathbf{H}=-\sum_{i=1}^{n}\mathbf{B}_i \hat{\mathbf{r}}_i\\
&\mathbf{a}_i=\mathbf{A}_i(\hat{\mathbf{u}}\mathbf{p}^b_{is}+\hat{\mathbf{u}}\dot{\mathbf{r}}_i+\ddot{\mathbf{r}}_i),\mathbf{b}_i=\hat{\mathbf{r}}_i\mathbf{a}_i\\
&\mathbf{a}=\sum_{i=1}^{n}\mathbf{a}_i,\mathbf{b}=\sum_{i=1}^{n}\mathbf{b}_i\\
&\mathbf{p}^b_{is}=\hat{\mathbf{u}}\mathbf{r}_i+\dot{\mathbf{r}}_i+\mathbf{v}\\
&\mathbf{p}^b_{iss}=\hat{\mathbf{u}}\mathbf{p}^b_{is}+\hat{\dot{\mathbf{u}}}\mathbf{r}_i+\hat{\mathbf{u}}\dot{\mathbf{r}}_i+\ddot{\mathbf{r}}_i+\dot{\mathbf{v}}\\
\end{aligned}
\right.
\end{equation}

\noindent where ${\mathbf{r}}_i$ and $T_i$ show every tendon’s offset from the cross-section center of mass and every tendon’s tension, respectively. Also, $"b"$ denotes the body coordinate frame. By substituting equation (\ref{f_tendon}) into equation (\ref{f_and_l}), and then substituting the resulting equation into equation (\ref{PDE}), the derivatives of the internal force and moment with respect to $s$ can be expressed as:

\begin{equation}\label{ns}
\left\{
\begin{aligned}
&\mathbf{n}_s={\rho A\mathbf{R}}(\hat{\boldsymbol{\omega}}\mathbf{q}+\mathbf{q}_t)-\mathbf{R}(\mathbf{a}+\mathbf{A}\mathbf{v}_s+\mathbf{G}\mathbf{u}_s)-{\rho A\mathbf{g}}+\\
&\mathbf{RCq}\odot|\mathbf{q}|\\
&\mathbf{m}_s=({\rho \mathbf{RJ}}\boldsymbol{\omega})_t-\hat{\mathbf{p}}_s \mathbf{n} -\mathbf{R}(\mathbf{b}+\mathbf{B}\mathbf{v}_s+\mathbf{H}\mathbf{u}_s)-\mathbf{l}_e
\end{aligned}
\right.
\end{equation}

The dynamic system, which is shown in (\ref{PDE}) and (\ref{ns}), has independent variables $\mathbf{u}$ and $\mathbf{v}$, and as such, the linear elasticity law with material damping can be used to connect $\mathbf{u}$ and $\mathbf{v}$ to the internal loading as follows:

\begin{equation}\label{n_and_m}
\left\{
\begin{aligned}
&\mathbf{n}=\mathbf{R}[\mathbf{K}_{\mathbf{se}} (\mathbf{v}-\mathbf{v}^*)+\mathbf{B}_{\mathbf{se}}\mathbf{v}_t]\\
&\mathbf{m}=\mathbf{R}[\mathbf{K}_{\mathbf{bt}} (\mathbf{u}-\mathbf{u}^*)+\mathbf{B}_{\mathbf{bt}}\mathbf{u}_t]
\end{aligned}
\right.
\end{equation}

\noindent where $\mathbf{v}^*$ and $\mathbf{u}^*$ are the initial posture of a rod that are defined as $\mathbf{v}^*=\begin{bmatrix}
    0 & 0& 1 \\
\end{bmatrix}$ and $\mathbf{u}^*=0$ for our simulation, which means the rod is straight in $z$ direction, initially. Also, $\mathbf{K_{se}}$, $\mathbf{K_{bt}}$, $\mathbf{B_{se}}$, and $\mathbf{B_{bt}}$  correspond to stiffness matrix for shear and extension, stiffness matrix for bending and torsion, viscous damping matrix for shear and extension, and viscous damping matrix for bending and torsion, respectively.
In (\ref{ns}), $\mathbf{u}_s$ and $\mathbf{v}_s$ are implicit, so it is preferred to use $\mathbf{u}$ and $\mathbf{v}$ as state variables, which makes solving the final equations easier. By differentiating (\ref{n_and_m}), using (\ref{ns}), and applying time discretization (as explained in the next section), the final equation is obtained \cite{till2019}:

\begin{equation}\label{final_vs}
\begin{aligned}
    \begin{bmatrix}
    \mathbf{v}_s \\
    \mathbf{u}_s
\end{bmatrix}=
\begin{bmatrix}
   \boldsymbol{\theta}_{11} &\boldsymbol{\theta}_{12} \\
    \boldsymbol{\theta}_{21} & \boldsymbol{\theta}_{22}
\end{bmatrix}^{-1}
&\begin{bmatrix}
    \boldsymbol{\prod} \mathbf{n}-\sum \mathbf{n} \\
    \boldsymbol{\prod} \mathbf{m}-\sum \mathbf{m}
\end{bmatrix}
\end{aligned}
\end{equation}

\noindent where $\boldsymbol{\theta}_{11}=\mathbf{K_{se}}+c_0\mathbf{B_{se}+A}$, $\boldsymbol{\theta}_{12}=\mathbf{G}$, $\boldsymbol{\theta}_{21}=\mathbf{B}$, $\boldsymbol{\theta}_{22}=\mathbf{K_{bt}}+c_0\mathbf{B_{bt}+H}$, and

\begin{equation*}\label{pai}
\left\{
\begin{aligned}
&\boldsymbol{\prod} \mathbf{n}={\rho A}(\hat{\boldsymbol{\omega}}\mathbf{q}+\mathbf{q}_t)-\mathbf{R}^T \rho A\mathbf{g}+\mathbf{Cq}\odot|\mathbf{q}|-\mathbf{a}\\\\
&\boldsymbol{\prod} \mathbf{m}=\mathbf{R}^T( \rho \mathbf{RJ}\boldsymbol{\omega})_t-\mathbf{R}^T \hat{\mathbf{p}}_s \mathbf{n} -\mathbf{R}^T \mathbf{l}_e-\mathbf{b}\\
\end{aligned}
\right.
\end{equation*}

\begin{equation*}\label{Rns_and_ms2}
\left\{
\begin{aligned}
&\sum \mathbf{n}=\hat{\mathbf{u}}\mathbf{K}_{\mathbf{se}} (\mathbf{v}-\mathbf{v}^*)+\hat{\mathbf{u}}\mathbf{B}_{\mathbf{se}} (c_0\mathbf{v}+\mathbf{v}^h)-...\\
&\mathbf{K}_{\mathbf{se}}{\mathbf{v}_s}^*+\mathbf{B}_{\mathbf{se}}{\mathbf{v}_s}^h\\
&\sum \mathbf{m}=\hat{\mathbf{u}}\mathbf{K}_{\mathbf{bt}} (\mathbf{u}-\mathbf{u}^*)+\hat{\mathbf{u}}\mathbf{B}_{\mathbf{bt}} (c_0\mathbf{u}+\mathbf{u}^h)-...\\
&\mathbf{K}_{\mathbf{bt}} {\mathbf{u}_s}^*+\mathbf{B}_{\mathbf{bt}}{\mathbf{u}_s}^h.\\
\end{aligned}
\right.
\end{equation*}

To complete the equations of motion, it is necessary to determine the boundary conditions (BCs). In this case, it is assumed that there are tendons attached to the free end. Due to this, the boundary conditions at the fixed end of the robot and the free end of it are as follows:

At fixed end, we have: 

\begin{equation}\label{s=0}
{s}=0\rightarrow
\mathbf{p}=\mathbf{q}=\boldsymbol{\omega}=\mathbf{0},\ \mathbf{R}=\mathbf{I}_{3\times3}
\end{equation}

At free end:

\begin{equation}\label{s=L}
{s}={L}\rightarrow
\mathbf{n}=-{T}_i\frac{\mathbf{p}_{is}(L^-)}{||\mathbf{p}_{is}(L^-)||}, \ \mathbf{m}=-\mathbf{r}_i\times \mathbf{n}
\end{equation}

\noindent where
$\mathbf{p}_{is}=\mathbf{R}(\hat{\mathbf{u}}\mathbf{r}_i+\dot{\mathbf{r}}_i+\mathbf{v}).$

\section{PDEs numerical solution}
\label{sec3}
To convert the PDEs to a set of Ordinary Differential Equations (ODEs), the method of BDF-$\alpha$ is used, which calculates the time derivation terms. This method is $O(\delta t^2)$ accurate for a timestep $\delta t$ that can be described by:

\begin{equation}\label{yt}
\mathbf{y}_t^{(i)} = c_0 \mathbf{y}^{(i)} + \mathbf{y}_h^{(i)}
\end{equation}

\noindent where $\mathbf{y}^{(i)}(s)=\mathbf{y}(t_i ,s)$ and $\mathbf{y}_h^{(i)}=c_1 \mathbf{y}^{(i-1)}+c_2 \mathbf{y}^{(i-2)}+d_1 \mathbf{y}_t^{(i-1)} $. The coefficients are:

\begin{equation}\label{c0}
\left\{
\begin{aligned}
&c_0=(1.5+\alpha)/[\delta t(1+\alpha)]\\
&c_1=-2/\delta t,\ d_1=\alpha/(1+\alpha)\\
&c_2=(0.5+\alpha)/[\delta t(1+\alpha)]\\
\end{aligned}
\right.
\end{equation}

The time derivative of (\ref{PDE}) can be represented as:

\begin{equation}\label{dot_t}
\left\{
\begin{aligned}
&\mathbf{v}_t=c_0\mathbf{v}+\mathbf{v}_h,\ \mathbf{u}_t=c_0\mathbf{u}+\mathbf{u}_h\\
&\mathbf{q}_t=c_0\mathbf{q}+\mathbf{q}_h,\ \boldsymbol{\omega}_t=c_0\boldsymbol{\omega}+\boldsymbol{\omega}_h\\
\end{aligned}
\right.
\end{equation}

The CR is divided into $N$ nodes, starting with $s=0$ and increasing to $s=L$. In order to solve the ODEs, the Forward Euler method is applied to every node as follows:

\begin{equation}\label{ODE}
\left\{
\begin{aligned}
&\mathbf{p}(i,j+1)=\mathbf{p}(i,j)+\mathbf{p}_s ds,\\
&\mathbf{R}(i,j+1)=\mathbf{R}(i,j)+\mathbf{R}_s ds\\
&\mathbf{n}(i,j+1)=\mathbf{n}(i,j)+\mathbf{n}_s ds,\\
&\mathbf{m}(i,j+1)=\mathbf{m}(i,j)+\mathbf{m}_s ds\\
&\mathbf{q}(i,j+1)=\mathbf{q}(i,j)+\mathbf{q}_s ds,\\
&\boldsymbol{\omega}(i,j+1)=\boldsymbol{\omega}(i,j)+\boldsymbol{\omega}_s ds
\end{aligned}
\right.
\end{equation}

In this case, $i$ and $j$ represent the interval of time and space, respectively, and $ds$ represents the distance between two nodes. 

In every time interval, by having:

$\mathbf{X}=[\mathbf{p}(i,1) \; \mathbf{R}(i,1) \; \mathbf{n}(i,1) \; \mathbf{m}(i,1) \; \mathbf{q}(i,1) \; \boldsymbol{\omega}(i,1)]$, and using (\ref{PDE}), (\ref{f_tendon}), and (\ref{final_vs}) to calculate $[\mathbf{p}_s \quad \mathbf{R}_s \quad \mathbf{n}_s \quad \mathbf{m}_s \quad \mathbf{q}_s \quad \boldsymbol{\omega}_s]
$, the states of the next node can be calculated. When the boundary conditions (states in the first node, $s=0$) are known, the state at the end of the CR can be determined using (\ref{ODE}). Based on (\ref{s=0}) and (\ref{s=L}), the force and moment ($\mathbf{n}$ and $\mathbf{m}$) at the first node are unknown, so the shooting method is used. The method consists of guessing the unknown boundary conditions $\mathbf{n}(i,1)$ and $\mathbf{m}(i,1)$, and updating those guessed values through an optimization routine so that the residual error of $\mathbf{n}$ and $\mathbf{m}$ at the free end approaches zero.
As a result, we summarize the process of solving the CR's dynamic equation in Algorithm 1 as follows:

\begin{algorithm}
\caption{Dynamic Simulation of a TDCR }
\label{alg1}
\textbf{Input:} The initial conditions (ICs), the boundary conditions (BCs), physical and mechanical properties, and the tendons’ tension.\\
\\
\textbf{for} $t = 0 : t$ \textbf{do:}\\
\hspace*{1.5em} fsolve $\rightarrow$  Shooting method to minimize the error of the BCs at free end\\
\hspace*{1.5em} \textbf{for} $s = 0 : L$ \textbf{do:}\\
\hspace*{3em} Calculate implicit approximation of time derivatives (\ref{yt})\\
\hspace*{3em} Implement spatial integration using forward Euler's method (\ref{ODE})\\
\hspace*{1.5em} \textbf{end for}\\
\textbf{end for}\\
Visualize the dynamic response over $t$
\end{algorithm}

\section{Control Design}
\label{sec4}
In this section, sliding mode control, which is commonly used in the literature, is formulated for a TDCR, and next, our backstepping-based technique is presented.

\subsection{Sliding mode control design}
This section focuses on how sliding mode control can be used to control the end position of a TDCR under planar motion \cite{alqumsan2019a,alqumsan2019b,alqumsan2021,samadikhoshkho2020,tang2023}. The first and second state variables, representing the position and velocity of the robot’s end-effector, are defined as:

\begin{equation}\label{SMC1}
\mathbf{X}_1=\mathbf{p}_{\text{end}}
\end{equation}

\begin{equation}\label{SMC2}
\mathbf{X}_2=\dot{\mathbf{X}}_1=\mathbf{V}_{\text{end}}=(\mathbf{p}_t)_{\text{end}}=(\mathbf{Rq})_{\text{end}}
\end{equation}

Based on the dynamics of the TDCR, we derive \(\dot{\mathbf{X}}_2\), resulting in the state-space model:

\begin{equation}\label{stateSpace}
\left\{
\begin{aligned}
&\dot{\mathbf{X}}_1=\mathbf{X}_2\\
&\dot{\mathbf{X}}_2=\frac{1}{{\rho A}} (\mathbf{n}_s + \mathbf{f}_e + \mathbf{f}_{\text{tendon}})_{\text{end}}\\
\end{aligned}
\right.
\end{equation}

\noindent Here, \(\mathbf{f}_{\text{tendon}}\) can be expressed as:

\begin{equation}\label{SMC5}
\mathbf{f}_{\text{tendon}}=
-\begin{bmatrix}
    \boldsymbol{\alpha}_1 & \dots  & \boldsymbol{\alpha}_n \\
\end{bmatrix}_{3\times n}
\begin{bmatrix}
    {T}_1 \\
    \vdots\\
    {T}_n
\end{bmatrix}_{n\times 1}=-\boldsymbol{\alpha}_M \mathbf{T}_M
\end{equation}

\noindent where \(\boldsymbol{\alpha}_i=\frac{(\hat{\mathbf{p}}_{is})^2}{||\mathbf{p}_{is}||^3} \mathbf{p}_{i_{ss}}\).

\vspace{0.25cm}

The final form of the state-space model is:

\begin{equation}\label{stateSpace_final}
\left\{
\begin{aligned}
&\dot{\mathbf{X}}_1=\mathbf{X}_2\\
&\dot{\mathbf{X}}_2=\mathbf{a}_c+\mathbf{b}_c\mathbf{U}
\end{aligned}
\right.
\end{equation}

\noindent where \(\mathbf{a}_c=\frac{1}{{\rho A}} (\mathbf{n}_s + \mathbf{f}_e)_{\text{end}}, \mathbf{b}_c=\frac{-1}{{\rho A}} (\boldsymbol{\alpha}_M)_{\text{end}}\), and \(\mathbf{U}=\mathbf{T}_M\).

\vspace{0.5cm}

To design the sliding mode control, we define a sliding mode surface \(\mathbf{S}\) as a linear combination of the position error \((\mathbf{e} = \mathbf{X}_d - \mathbf{X}_1)\) and its derivative \((\dot{\mathbf{e}} = \dot{\mathbf{X}}_d - \dot{\mathbf{X}}_1)\). \(\mathbf{S}\) is designed to ensure rapid response and stability. As \(\mathbf{S} \rightarrow 0\), both \(\mathbf{e}\) and \(\dot{\mathbf{e}}\) converge to zero, aligning the system state with the desired trajectory. The sliding mode surface \(\mathbf{S}\) is defined as:

\begin{equation}\label{S}
\mathbf{S} = \dot{\mathbf{e}} + c\mathbf{e}, \quad c > 0
\end{equation}

Applying the exponential reaching law \(\left(\dot{\mathbf{S}} = -\varepsilon \, \text{sgn}(\mathbf{S}) - k\mathbf{S}, \, \varepsilon, k > 0\right)\), the control input is formulated as:

\begin{equation}\label{Sdot3}
\mathbf{U} = \frac{1}{\mathbf{b}_c} \left[c(\dot{\mathbf{X}}_d - \dot{\mathbf{X}}_1) + \ddot{\mathbf{X}}_d - \mathbf{a}_c + \varepsilon \, \text{sgn}(\mathbf{S}) + k\mathbf{S} \right].
\end{equation}

By defining the Lyapunov function as \(\mathbf{V} = \frac{1}{2} \mathbf{S}^2\), the stability condition is shown as:

\begin{equation}\label{Sdot66}
\dot{\mathbf{V}} = \mathbf{S}\dot{\mathbf{S}}(t) = -k\mathbf{S}^2 - \varepsilon|\mathbf{S}| \leq -2k\mathbf{V} \longrightarrow \mathbf{V}(t) \leq e^{-2kt} \mathbf{V}(0)
\end{equation}

From (\ref{Sdot66}), we infer that the sliding mode surface \(\mathbf{S}\) tends to zero exponentially as \(k\) increases.

\subsection{Backstepping control design}
For the design of a backstepping control, it is necessary to express the system in the form of $\dot{\mathbf{X}} = f(\mathbf{X}, \mathbf{U})$, where $\mathbf{X} = \begin{pmatrix} \mathbf{x}_1 \ \mathbf{x}_2 \end{pmatrix}^T$. This yields:

\begin{equation}\label{Back1}
f(\mathbf{X}, \mathbf{U}) =
  \begin{bmatrix}
    \mathbf{x}_2 \\
    \mathbf{a}_c + \mathbf{b}_c \mathbf{U} \\
  \end{bmatrix}  
\end{equation}

\noindent The tracking error is defined as:
\begin{equation}\label{Back2}
\mathbf{z}_1 = \mathbf{x}_{1d} - \mathbf{x}_1.
\end{equation}

\noindent The Lyapunov function is expressed as:

\begin{equation}\label{Back3}
\mathbf{V}(\mathbf{z}_1) = \frac{1}{2} \mathbf{z}_1^2.
\end{equation}

\noindent Its time derivative function should be negative semi-definite, indicating:

\begin{equation}\label{Back4}
\begin{aligned}
&\dot{\mathbf{V}}(\mathbf{z}_1) = \mathbf{z}_1 \dot{\mathbf{z}}_1 = \mathbf{z}_1(\dot{\mathbf{x}}_{1d} - \mathbf{x}_2) \stackrel{\text{for } \dot{\mathbf{V}}(\mathbf{z}_1) < 0}{\longrightarrow} \\
&\mathbf{x}_2 = \dot{\mathbf{x}}_{1d} + \alpha_1 \mathbf{z}_1 \quad \alpha_1 > 0 \rightarrow \ \dot{\mathbf{V}}(\mathbf{z}_1) = -\alpha_1 \mathbf{z}_1^2
\end{aligned}
\end{equation}

\noindent A new variable is introduced as:

\begin{equation}\label{Back5}
\begin{aligned}
&\mathbf{z}_2 = \mathbf{x}_2 - \dot{\mathbf{x}}_{1d} - \alpha_1 \mathbf{z}_1
\end{aligned}
\end{equation}

\noindent In the next step, the augmented Lyapunov function is considered as:

\begin{equation}\label{Back6}
\begin{aligned}
&\mathbf{V}(\mathbf{z}_1, \mathbf{z}_2) = \frac{1}{2}(\mathbf{z}_1^2 + \mathbf{z}_2^2)
\end{aligned}
\end{equation}

\noindent Its time derivative is given by:

\begin{equation}\label{Back7}
\begin{aligned}
&\dot{\mathbf{V}}(\mathbf{z}_1,\mathbf{z}_2) = \mathbf{z}_1\dot{\mathbf{z}}_1+\mathbf{z}_2\dot{\mathbf{z}}_2 \\
&= \mathbf{z}_1\dot{\mathbf{z}}_1+\mathbf{z}_2(\dot{\mathbf{x}}_2-\ddot{\mathbf{x}}_{1d}-\alpha_1\dot{\mathbf{z}}_1) \\
&= \mathbf{z}_2(\mathbf{a}_c+\mathbf{b}_c\mathbf{U})-\mathbf{z}_2\ddot{\mathbf{x}}_{1d}+\dot{\mathbf{z}}_1(\mathbf{z}_1-\mathbf{z}_2\alpha_1) \\
&= \mathbf{z}_2(\mathbf{a}_c+\mathbf{b}_c\mathbf{U})-\mathbf{z}_2\ddot{\mathbf{x}}_{1d}+(-\alpha_1\mathbf{z}_1-\mathbf{z}_2)(\mathbf{z}_1-\mathbf{z}_2\alpha_1) \\
&= \mathbf{z}_2(\mathbf{a}_c+\mathbf{b}_c\mathbf{U})-\mathbf{z}_2\ddot{\mathbf{x}}_{1d}-\alpha_1\mathbf{z}_1^2+\alpha_1^2\mathbf{z}_1\mathbf{z}_2-\mathbf{z}_1\mathbf{z}_2+\alpha_1\mathbf{z}_2^2
\end{aligned}
\end{equation}

The following amount of $\mathbf{U}$ satisfies the condition of $\dot{\mathbf{V}}<0$ .

\begin{equation}\label{Back8}
\begin{aligned}
&\mathbf{U} = \frac{1}{\mathbf{b}_c} (-\mathbf{a}_c + \mathbf{z}_1 - \alpha_1 \mathbf{z}_2 - {\alpha_1}^2 \mathbf{z}_1 - \alpha_2 \mathbf{z}_2 + \ddot{\mathbf{x}}_{1d})
\end{aligned}
\end{equation}

\noindent By substituting the value of $\mathbf{U}$ into $\dot{V}(\mathbf{z}_1, \mathbf{z}_2)$, the following is obtained:

\begin{equation}\label{Back9}
\begin{aligned}
\dot{\mathbf{V}}(\mathbf{z}_1,\mathbf{z}_2) = -\alpha_1 \mathbf{z}_1^2 - \alpha_2 \mathbf{z}_2^2 < 0 \quad \text{for} \quad \alpha_1, \alpha_2 > 0
\end{aligned}
\end{equation}

\noindent Therefore, the states of the system gradually decrease and tend towards zero as time progresses.

\section{Simulation and Experimental Results}
\label{sec5}
\begin{figure}[H]
  \centering
  \includegraphics[width=0.85\textwidth]{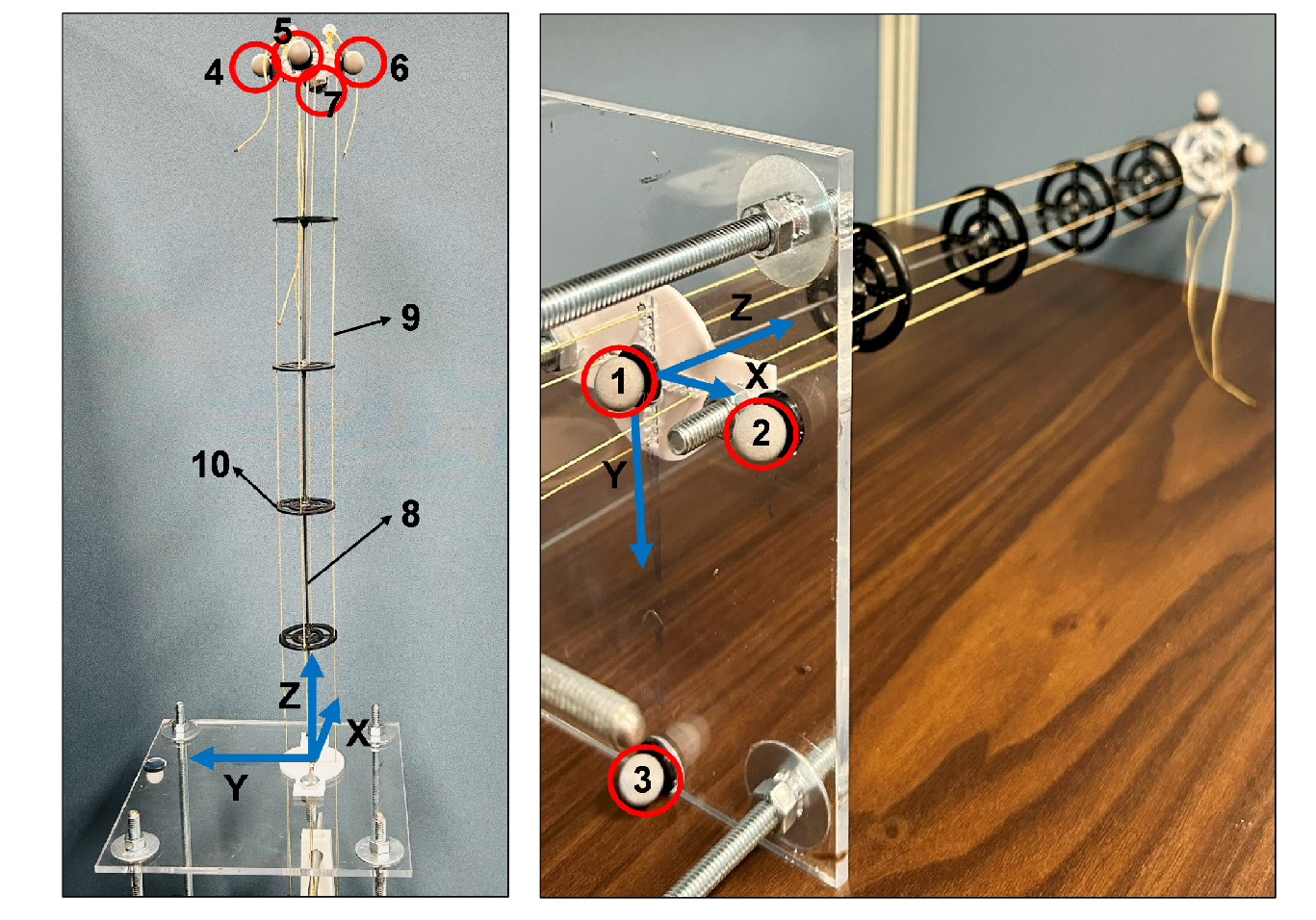}
  \caption{A tendon-driven continuum robot used in the test setup. Markers 1-3 indicate base markers, 4-7 denote CR's tip markers, 8 represents the backbone, 9 stands for tendon, and 10 indicates the spacer disk.}
  \label{Setup}
\end{figure}

Prior to detailing the simulation steps, it is essential to provide an overview of the experimental setup, as illustrated in Fig. \ref{Setup}. The single section CR's backbone consisted of 304 stainless steel with a length ($L$) of $0.5\ $m. Four Kevlar tendons were arranged around the backbone with a $2.0\ $cm offset and a $90$\textdegree$\ $angular separation. Spacer disks were 3D printed using PLA filament and attached to the backbone with epoxy glue to create a fixed and solid connection. Our prototype includes channels with tight tolerances for tendon routing, where the channel diameter closely matches the tendon diameter to minimize movement during deformation, supporting the second assumption in our model. The XL430-W250-T servomotors (Dynamixel, USA) were employed to actuate the robot during the test. Vicon motion system (Tracker 3.10, UK) was utilized to track the posture of the robot's end-effector in 3D space by attaching 7 markers (3 at the base and 4 at the tip of the CR). The structural values of the robot are detailed in Table \ref{tab111}.

\begin{table}[htbp]
\caption{Structural Values of the Continuum Robot}
\begin{center}
    \begin{tabular}{ccc}
        \toprule
        \textbf{Parameter} & \textbf{Value} & \textbf{Source} \\
        \midrule
        Density & $\rho=6,366$ $\mathrm{kg/m^3}$ & Data Sheet \\
        Young’s Modulus & $E= 190\ $ $\mathrm{Gpa}$ & Data Sheet \\
        Radius & $r=1\ $ $\mathrm{mm}$& Data Sheet \\
        Damping matrix \textbf{$\mathbf{B_{bt}}$} & $0.008 \times \mathbf{I}\ $ $\mathrm{N/m^2}$& Calibration\cite{till2019} \\
        Damping matrix \textbf{$\mathbf{B_{se}}$} & $0 $ & Calibration\cite{till2019} \\
        Damping matrix \textbf{$\mathbf{C}$} & $0.1 \times \mathbf{I}\ $  $\mathrm{kg/m^2}$& Calibration\cite{till2019} \\

        \bottomrule
    \end{tabular}

\label{tab111}
\end{center}
\end{table}

The numerical simulations involved implementing a $0.5 \ $m length steel rod as the CR, with material and geometrical properties detailed in Table \ref{tab111} . To account for the effects of glue, markers, and spacer disks, the mass density was recalculated to be $\rho= 17,189 \ $$\mathrm{kg/m^3}$. Additionally, a spatial resolution of $N = 200$ was assumed, and the time domain discretization parameters for BDF-$\alpha$ were adjusted to $\alpha=-0.2$ and  $dt=0.01$. The robot’s backbone initial position is laying on the $z$ axis, with $\mathbf{v^*}=[0\ 0\ 1]^T$ and $\mathbf{u^*}=0$. The damping coefficients for the arm are provided in Table \ref{tab111}.

The designed controllers are intended to drive the robot's end-effector to a defined reference signal, \(\mathbf{X}_d\), achieving a 90-degree orientation aligned with the x-axis. To accomplish this, \(\mathbf{X}_d\) is set to \([340 (1-e^{-20t})\ 0\ 0]^T \) mm, guiding the robot's tip to the desired position. The parameters for the sliding mode and backstepping controllers were chosen to achieve zero steady-state error, ensuring precise trajectory tracking and enhanced stability. This approach allows the system to reach the target position without residual error. The sliding mode and backstepping
controllers' parameters are:

\begin{equation}\label{Control_para}
\left\{
\begin{aligned}
&c=2100,\; k=12 \; , \mathrm{and} \; \; \varepsilon=0.005 \quad \mathrm{for \ SMC}\\
&\alpha_1=1500 \; \mathrm{and} \; \; \alpha_2=12.5 \quad \mathrm{for\;\; Backstepping}\\
\end{aligned}
\right.
\end{equation}

\begin{figure}[H]
    \centering
    \begin{subfigure}[b]{0.49\textwidth}
        \centering
        \includegraphics[width=\textwidth, trim={15 0 30 10}, clip]{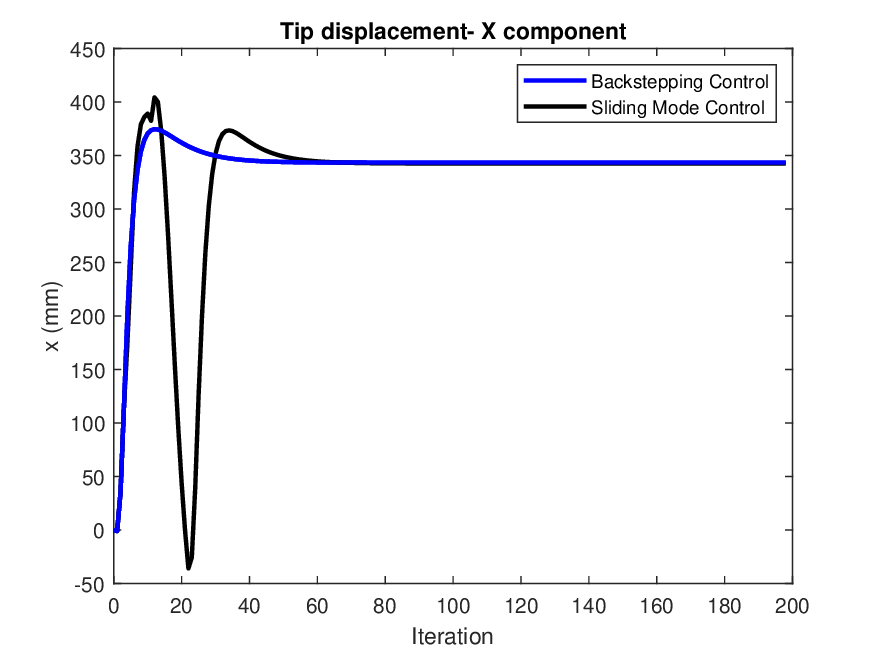}
        \caption{}
    \end{subfigure}
    \hfill
    \begin{subfigure}[b]{0.49\textwidth}
        \centering
        \includegraphics[width=\textwidth, trim={15 0 30 10}, clip]{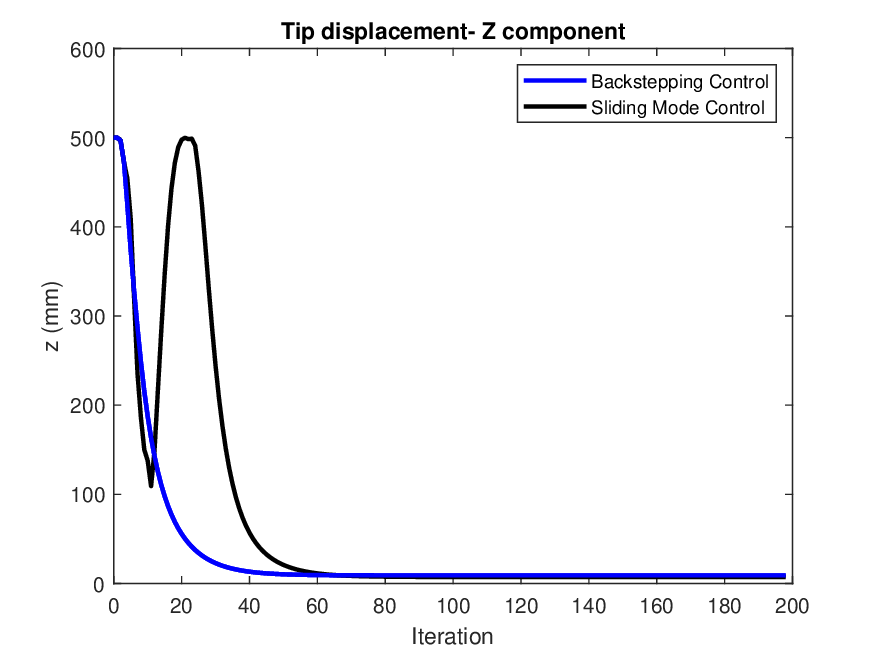}
        \caption{}
    \end{subfigure}
    \\
    \vspace{1em}
    \begin{subfigure}[b]{0.49\textwidth}
        \centering
        \includegraphics[width=\linewidth, trim={15 0 30 10}, clip]{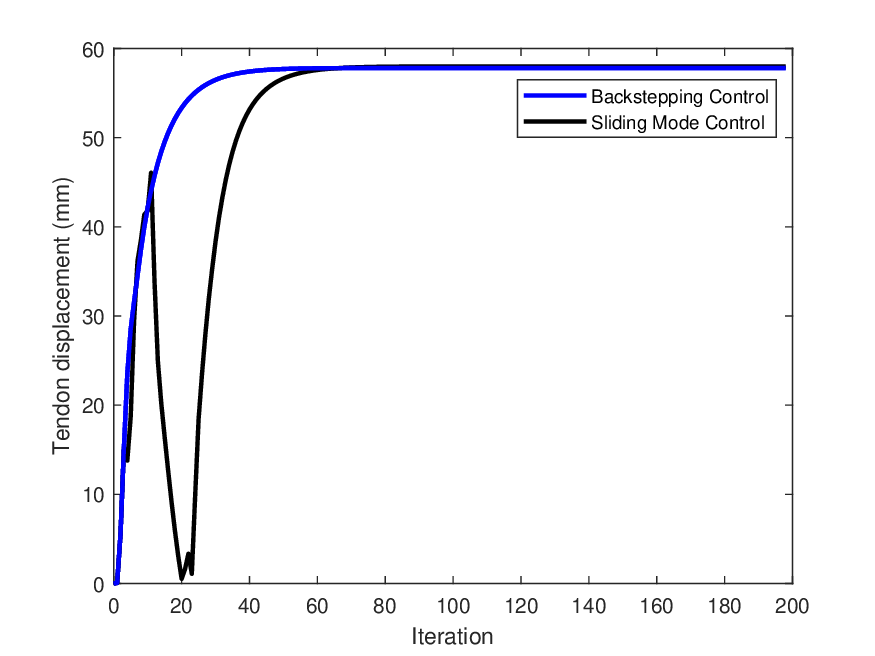}
        \caption{}
    \end{subfigure}
    \hfill
    \begin{subfigure}[b]{0.49\textwidth}
        \centering
        \includegraphics[width=\textwidth, trim={15 0 30 10}, clip]{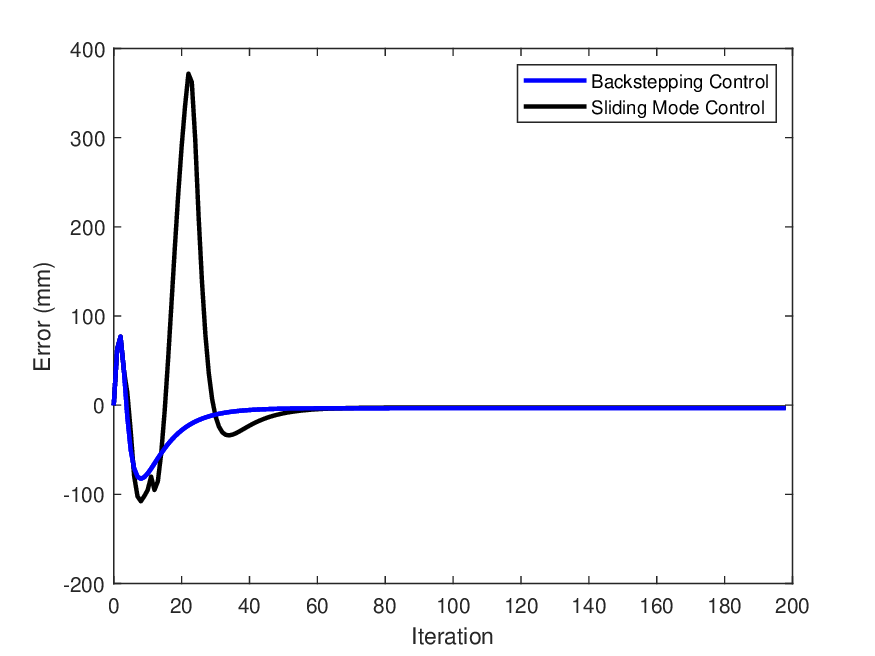}
        \caption{}
    \end{subfigure}
    \caption{Comparison of backstepping and sliding mode control under normal conditions: (a) x-component of the robot's tip vs. iteration, (b) z-component of the robot's tip vs. iteration, (c) tendon displacement, and (d) position error.}

    \label{Simulation_normal}
\end{figure}

Fig. \ref{Simulation_normal} (a) and (b) show the $x$ and $z$ components of the robot's tip, respectively. The $x$ component should reach $340 \ $mm according to the controllers' desired position, while the $z$ component starts at the point of $500 \ $mm (the length of the robot) and approaches zero. This indicates that the robot reaches a $90$\textdegree$\ $angle and experiences a large deflection. Both backstepping and sliding mode control methods can drive a robot's end-effector to a desired point, but they exhibit different behaviors. Backstepping enables a smooth approach, while sliding mode control results in drastic fluctuations.

Fig. \ref{Simulation_normal} (c) displays the tendon displacement, which serves as the input of the motors in the experimental setup. When backstepping control is employed, the tendon displacement starts at zero and gradually increases to $58\ $mm at the end. However, the use of sliding mode control results in undesirable ups and downs in tendon displacement, which is not ideal for the motor.

Fig. \ref{Simulation_normal} (d) shows the error signal for both controllers, which converges to zero, indicating that both controllers successfully drive the robot to the desired position.

Table \ref{ComparisonTable1} presents a quantitative comparison of backstepping control and sliding mode control under normal conditions, highlighting key performance metrics. The Total Path Length (TPL) serves as an indicator of smoothness, with backstepping control achieving a significantly shorter path length of 407.79 mm compared to 1300.12 mm for sliding mode control, highlighting the smoother trajectory of backstepping. In terms of settling time, backstepping control stabilizes the robot within 22 iterations, faster than the 43 iterations required by sliding mode control. The overshoot values also demonstrate a notable difference, with backstepping showing a lower overshoot of 9.1\% versus 18.2\% for sliding mode control, indicating more effective control over peak response. Both controllers achieve a rise time of 7 iterations from 0 to 90\%, but backstepping shows a distinct advantage in managing overshoot and achieving faster stabilization. Finally, both methods reach zero steady-state error, demonstrating accurate long-term tracking performance. Overall, the results suggest that backstepping control offers superior performance in terms of smoothness, settling time, and overshoot.

\begin{table}[h!]
\centering
\caption{Comparison of backstepping and sliding mode control under normal conditions.}
\label{ComparisonTable1}
\begin{tabular}{lcc}
\hline
\textbf{Criteria} & \textbf{Backstepping Control} & \textbf{Sliding Mode Control}\\
\hline
Smoothness & TPL= 407.79 mm & TPL= 1300.12 mm \\
Settling Time (5\%) & 22 iterations & 43 iterations \\
Overshoot & 9.1 \%  & 18.2 \% \\
Rise Time (0-90\%) & 7 iterations & 7 iterations \\
Steady-State Error & 0 & 0 \\
\hline
\end{tabular}
\end{table}

\begin{figure}[H]
  \centering
  \begin{subfigure}[b]{0.49\textwidth}
      \centering
      \includegraphics[width=\textwidth]{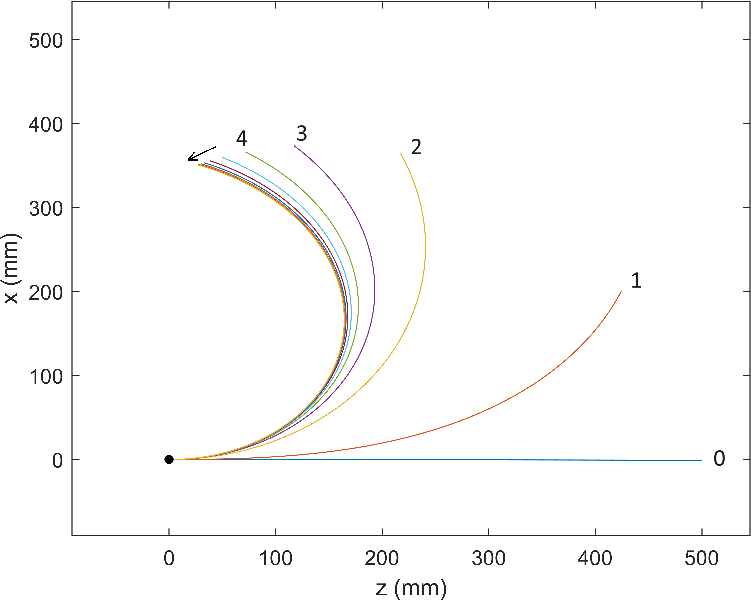}
      \caption{}
      
  \end{subfigure}
  \hfill
  \begin{subfigure}[b]{0.49\textwidth}
      \centering
      \includegraphics[width=\textwidth]{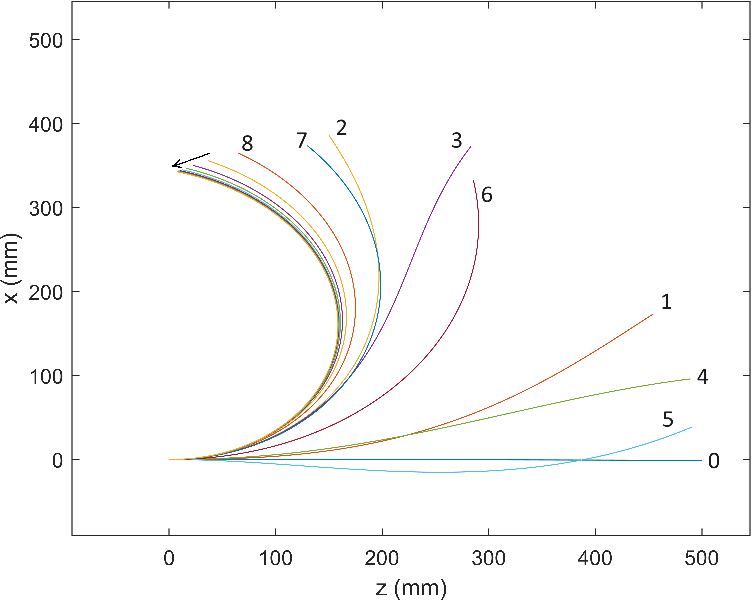}
      \caption{}
      
  \end{subfigure}
  \caption{Comparison of the robot's configuration under different control methods, (a) backstepping control, and (b) sliding mode control.}
  \label{CR_configurations}
\end{figure}

To better understand the performance of the two controllers, Fig. \ref{CR_configurations} illustrates the simulated TDCR configurations under backstepping and sliding mode control. The configuration of the CR is plotted every five iterations, beginning from its initial position lying on the z-axis, as shown in the zero configuration. With backstepping control, the tip of the robot gradually approaches the desired position in a smooth manner. In contrast, when using sliding mode control, the robot is first moved upwards, then downwards, before finally approaching the desired point.

To further validate the performance and robustness of the controllers, we examined two additional scenarios: external forces and disturbances.

In Scenario 1, an external force was introduced by adding weights (20 g and 50 g) to the robot's tip in the $-x$ direction, aligned with the gravitational force. The objective was to simulate the effect of a static load on the robot's end-effector, requiring the controllers to compensate for the added force.

\begin{figure}[H]
    \centering
    \begin{subfigure}[b]{0.49\textwidth}
        \centering
        \includegraphics[width=\textwidth, trim={15 0 30 10}, clip]{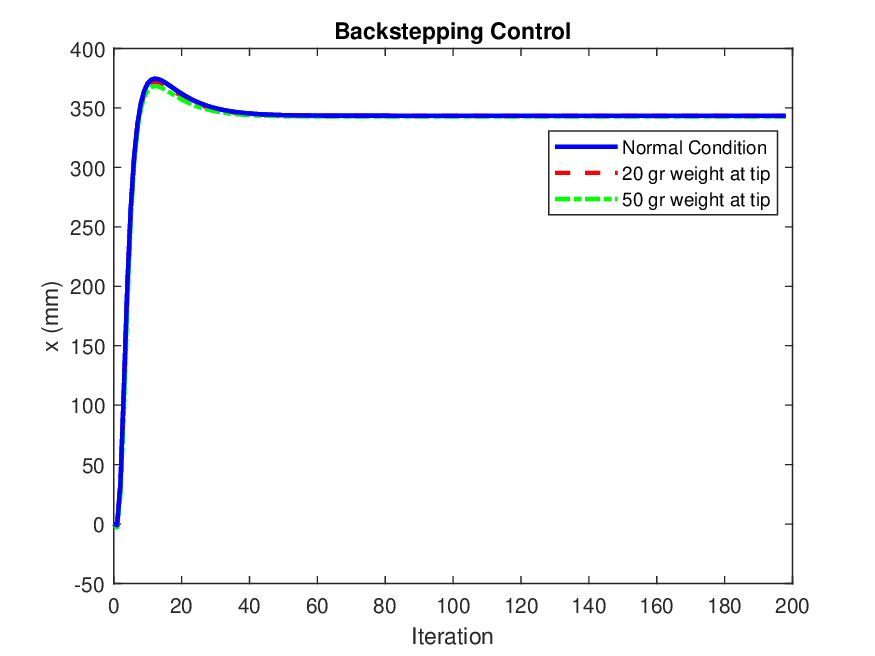}
        \caption{}
    \end{subfigure}
    \hfill
    \begin{subfigure}[b]{0.49\textwidth}
        \centering
        \includegraphics[width=\textwidth, trim={15 0 30 10}, clip]{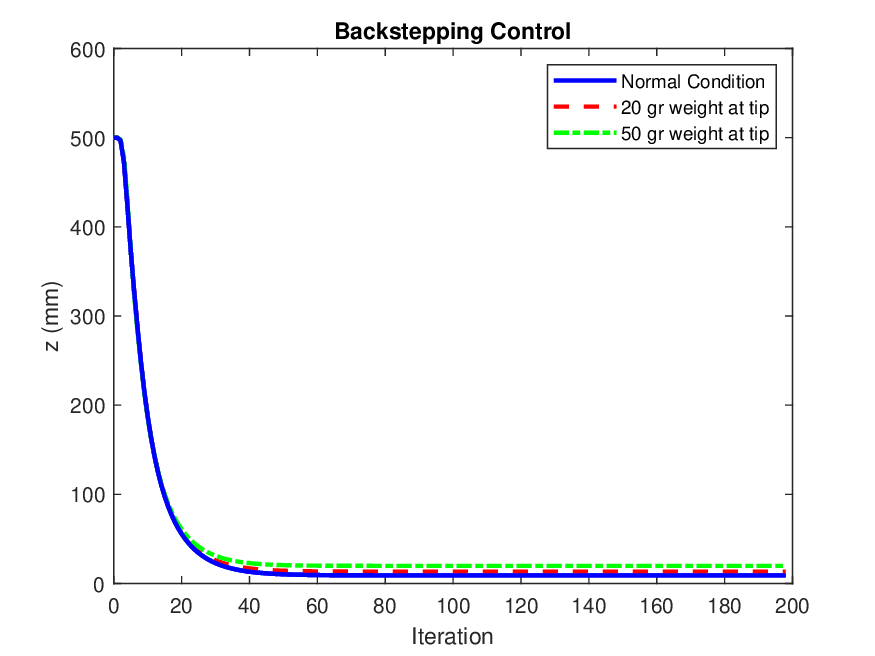}
        \caption{}
    \end{subfigure}
    \\
    \vspace{1em}
    \begin{subfigure}[b]{0.49\textwidth}
        \centering
        \includegraphics[width=\linewidth, trim={15 0 30 10}, clip]{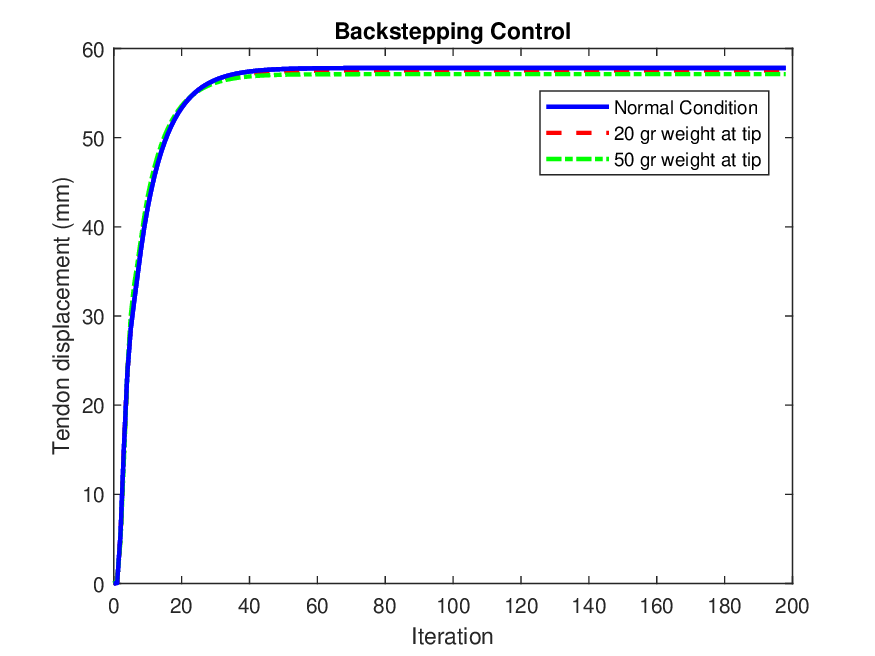}
        \caption{}
    \end{subfigure}
    \hfill
    \begin{subfigure}[b]{0.49\textwidth}
        \centering
        \includegraphics[width=\textwidth, trim={15 0 30 10}, clip]{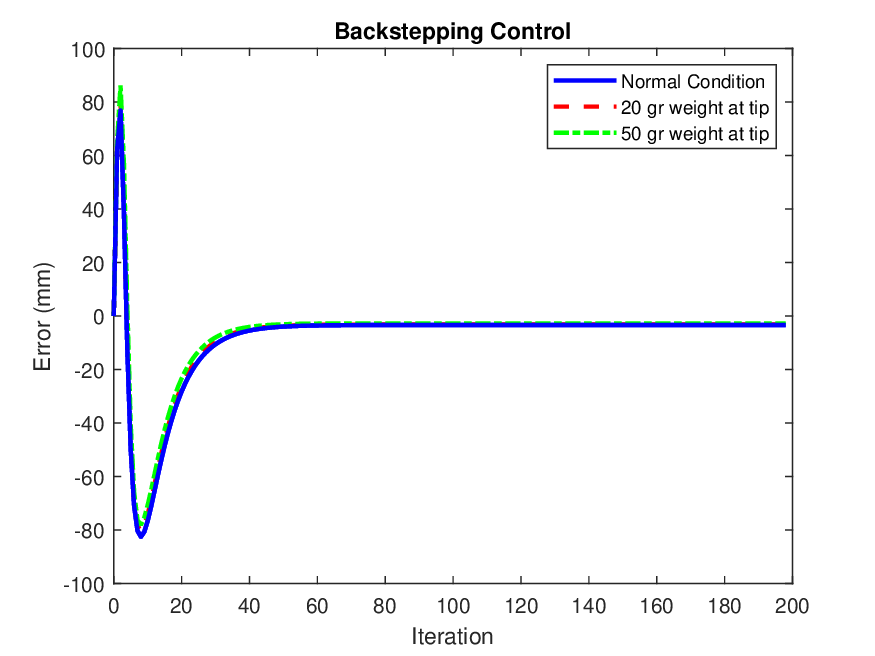}
        \caption{}
    \end{subfigure}
    \caption{ Backstepping control performance with added external weights: (a) x-component of the robot's tip vs. iteration, (b) z-component of the robot's tip vs. iteration, (c) tendon displacement, and (d) position error.
}
    \label{Simulation_BACK_Weight}
\end{figure}

\begin{figure}[H]
    \centering
    \begin{subfigure}[b]{0.49\textwidth}
        \centering
        \includegraphics[width=\textwidth, trim={15 0 30 10}, clip]{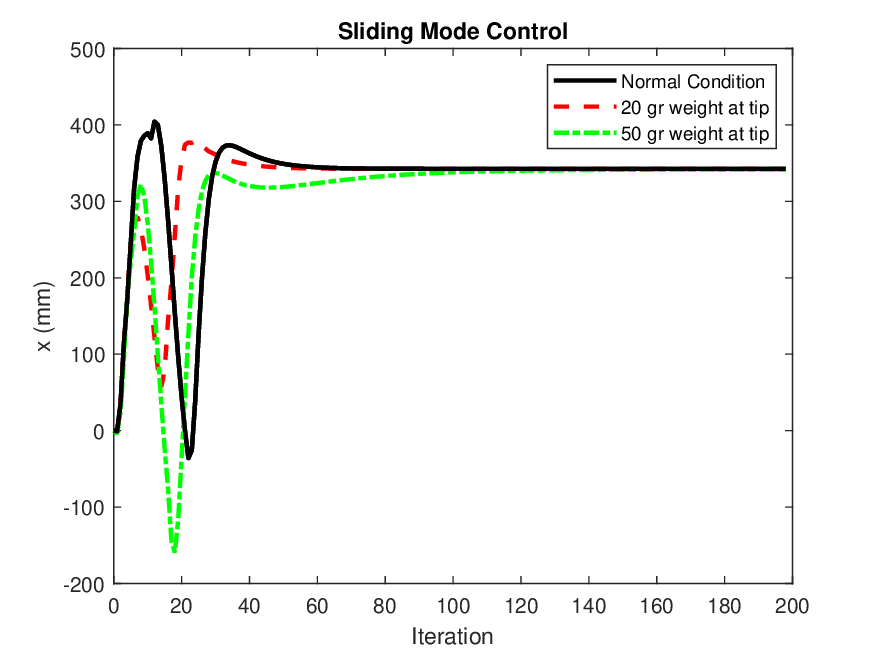}
        \caption{}
    \end{subfigure}
    \hfill
    \begin{subfigure}[b]{0.49\textwidth}
        \centering
        \includegraphics[width=\textwidth, trim={15 0 30 10}, clip]{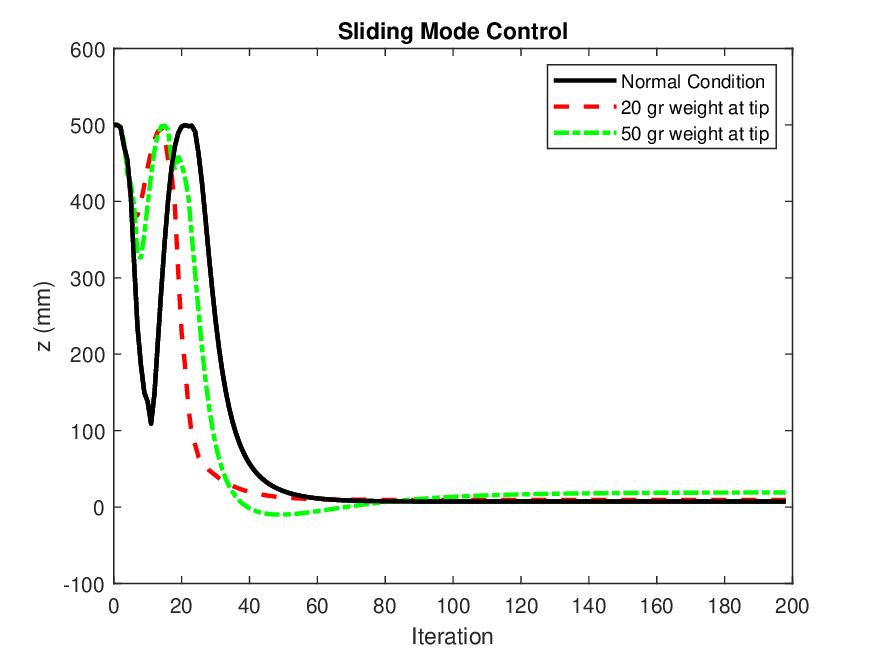}
        \caption{}
    \end{subfigure}
    \\
    \vspace{1em}
    \begin{subfigure}[b]{0.49\textwidth}
        \centering
        \includegraphics[width=\linewidth, trim={15 0 30 10}, clip]{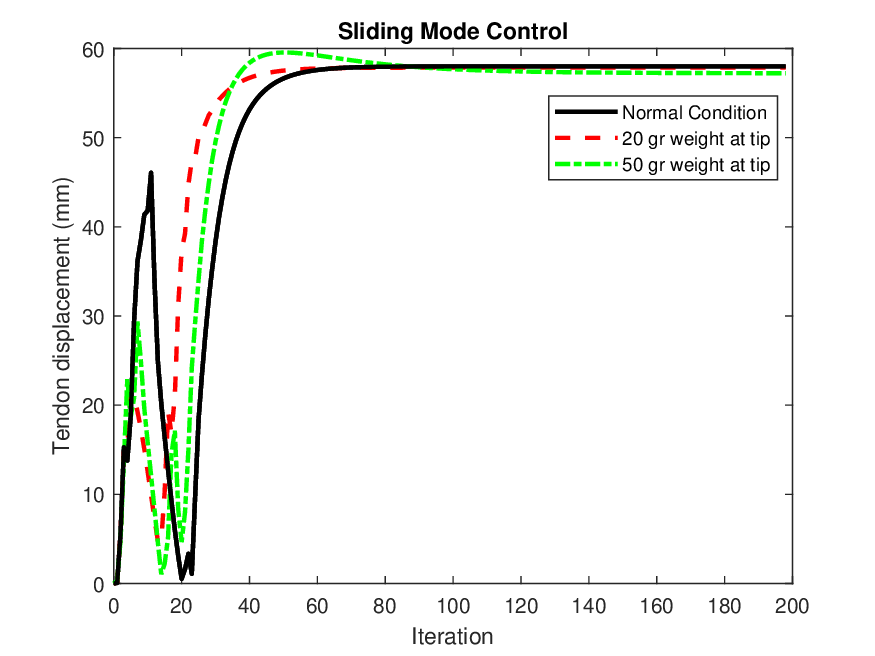}
        \caption{}
    \end{subfigure}
    \hfill
    \begin{subfigure}[b]{0.49\textwidth}
        \centering
        \includegraphics[width=\textwidth, trim={15 0 30 10}, clip]{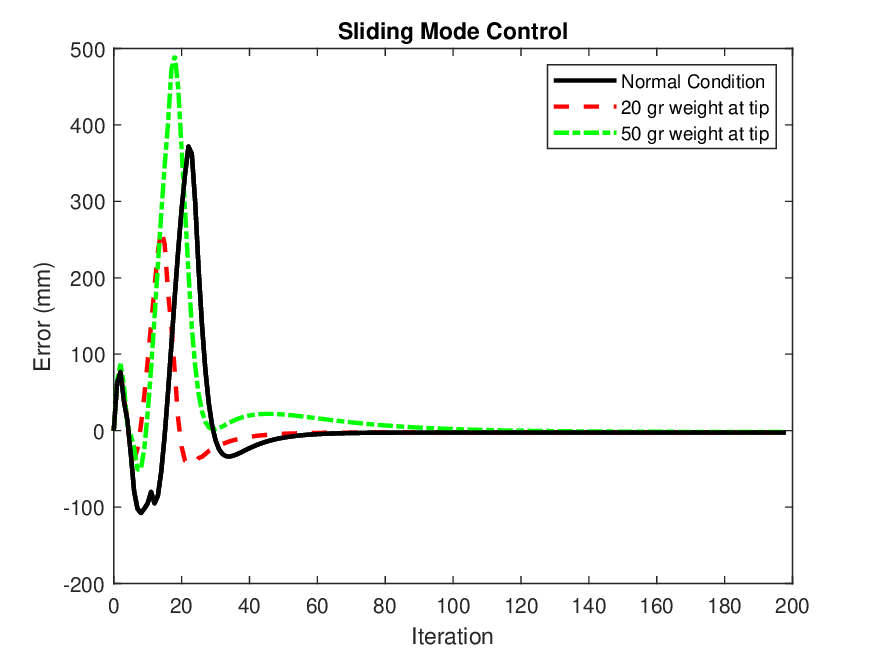}
        \caption{}
    \end{subfigure}
    \caption{Sliding mode control performance with added external weights: (a) x-component of the robot's tip vs. iteration, (b) z-component of the robot's tip vs. iteration, (c) tendon displacement, and (d) position error.}
    \label{Simulation_SMC_weight}
\end{figure}

Figs. \ref{Simulation_BACK_Weight} and \ref{Simulation_SMC_weight} illustrate the results of applying the external force. For backstepping control, the robot maintained a stable trajectory, smoothly adapting to the additional load without significant deviation. In contrast, the sliding mode control exhibited more fluctuations in response to the external weight, indicating a less stable response compared to backstepping control. These results highlight the robustness of the backstepping controller in handling external forces more effectively than sliding mode control.

In the second scenario, a disturbance force of \(10\ \text{N}\) was applied in the \(x\)-direction and \(-10\ \text{N}\) in the \(z\)-direction at iteration 50, challenging the controllers to promptly adjust and maintain system stability.

\begin{figure}[H]
    \centering
    \begin{subfigure}[b]{0.49\textwidth}
        \centering
        \includegraphics[width=\textwidth, trim={15 0 30 10}, clip]{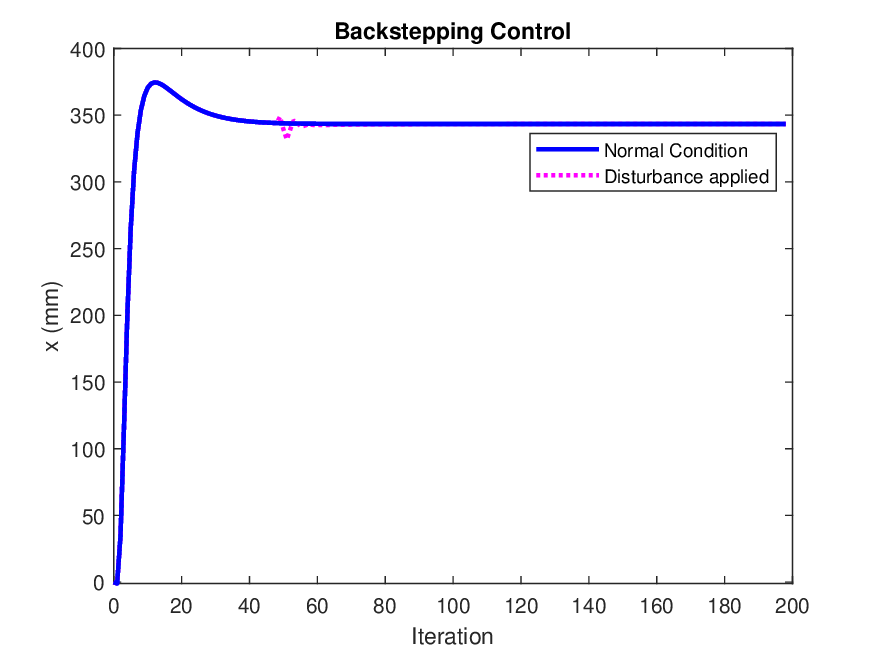}
        \caption{}
    \end{subfigure}
    \hfill
    \begin{subfigure}[b]{0.49\textwidth}
        \centering
        \includegraphics[width=\textwidth, trim={15 0 30 10}, clip]{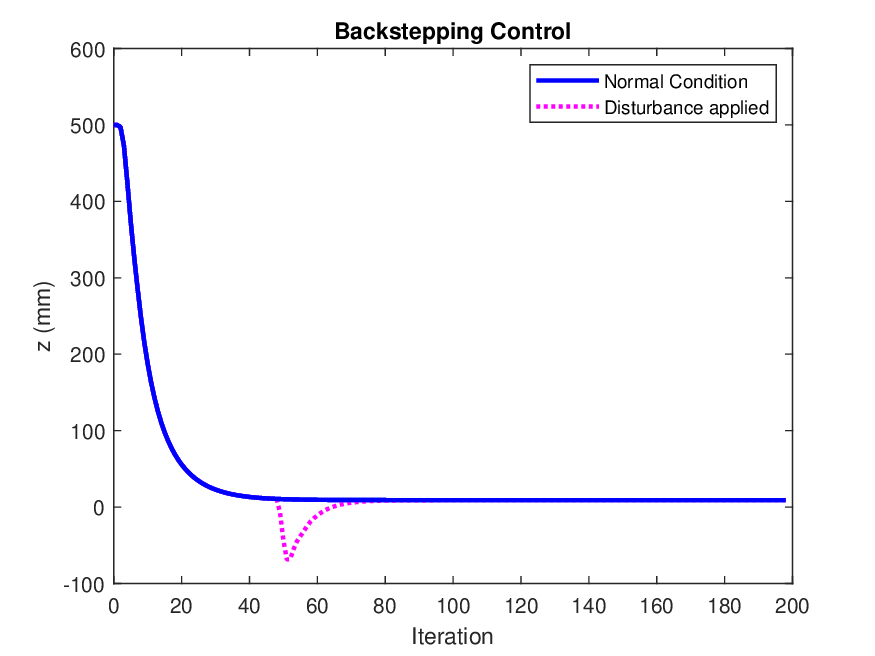}
        \caption{}
    \end{subfigure}
    \\
    \vspace{1em}
    \begin{subfigure}[b]{0.49\textwidth}
        \centering
        \includegraphics[width=\linewidth, trim={15 0 30 10}, clip]{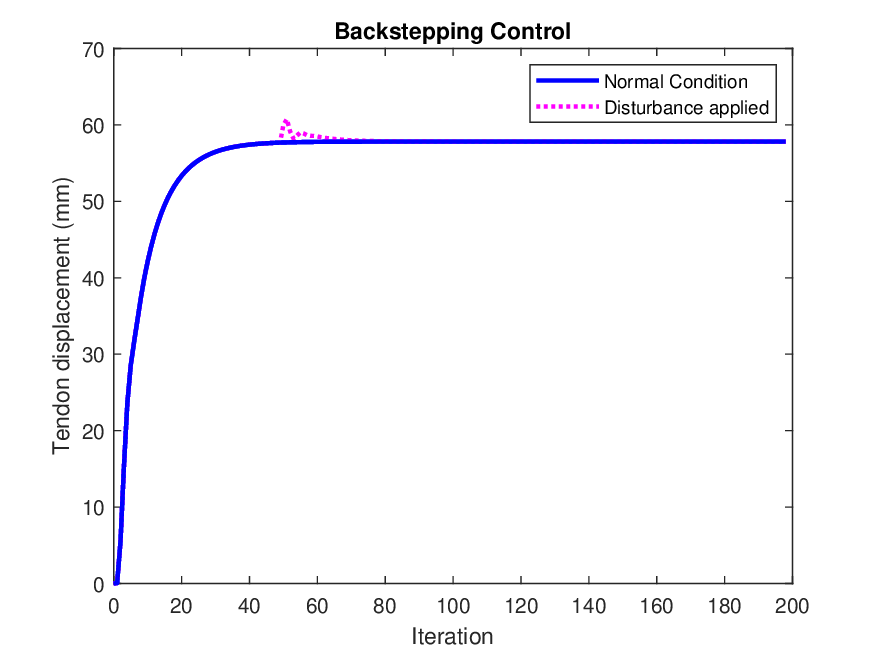}
        \caption{}
    \end{subfigure}
    \hfill
    \begin{subfigure}[b]{0.49\textwidth}
        \centering
        \includegraphics[width=\textwidth, trim={15 0 30 10}, clip]{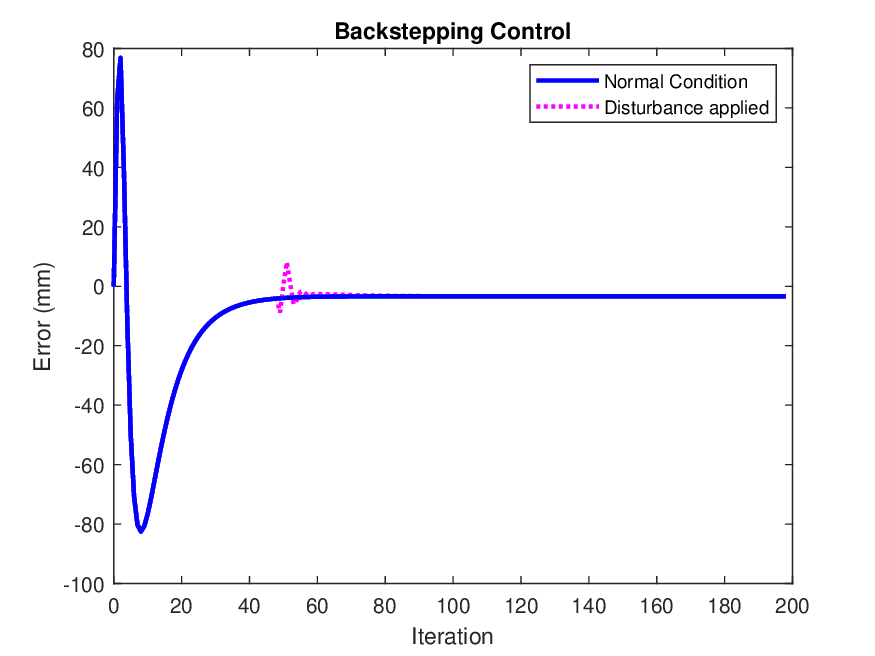}
        \caption{}
    \end{subfigure}
    \caption{Backstepping control performance under disturbance scenario: (a) x-component of the robot's tip vs. iteration, (b) z-component of the robot's tip vs. iteration, (c) tendon displacement, and (d) position error.
}
    \label{Simulation_BACK_Disturbance}
\end{figure}

\begin{figure}[H]
    \centering
    \begin{subfigure}[b]{0.49\textwidth}
        \centering
        \includegraphics[width=\textwidth, trim={15 0 30 10}, clip]{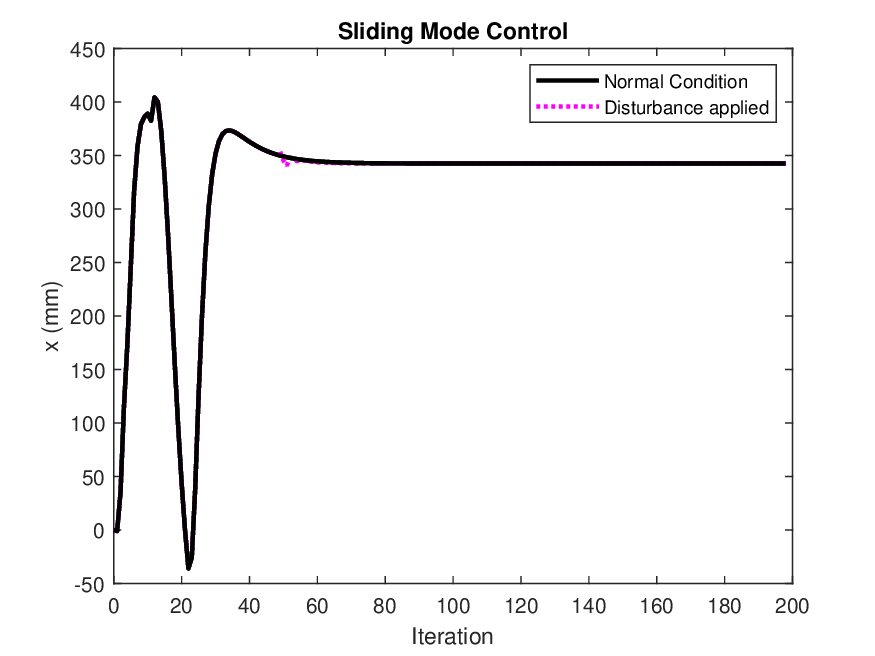}
        \caption{}
    \end{subfigure}
    \hfill
    \begin{subfigure}[b]{0.49\textwidth}
        \centering
        \includegraphics[width=\textwidth, trim={15 0 30 10}, clip]{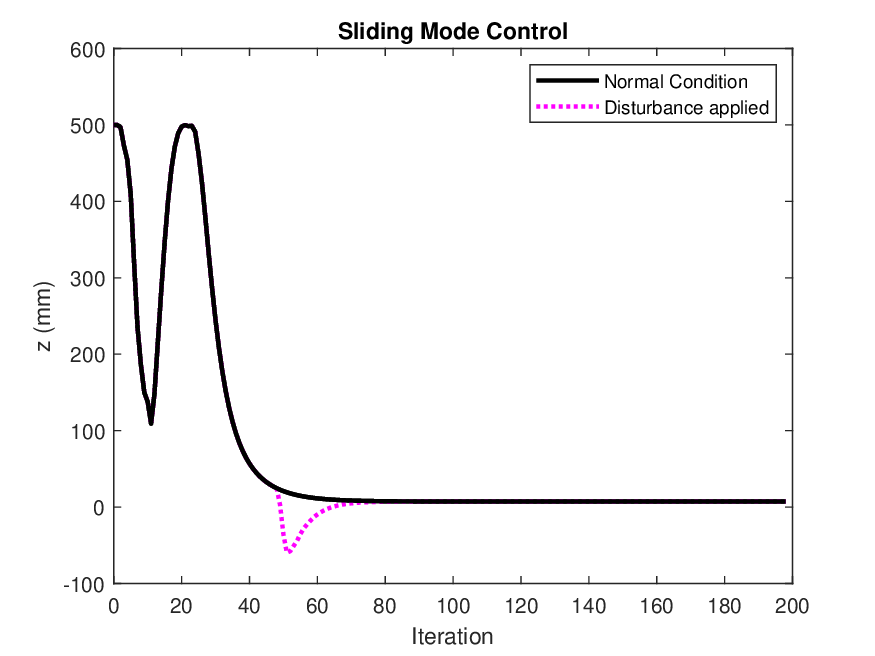}
        \caption{}
    \end{subfigure}
    \\
    \vspace{1em}
    \begin{subfigure}[b]{0.49\textwidth}
        \centering
        \includegraphics[width=\linewidth, trim={15 0 30 10}, clip]{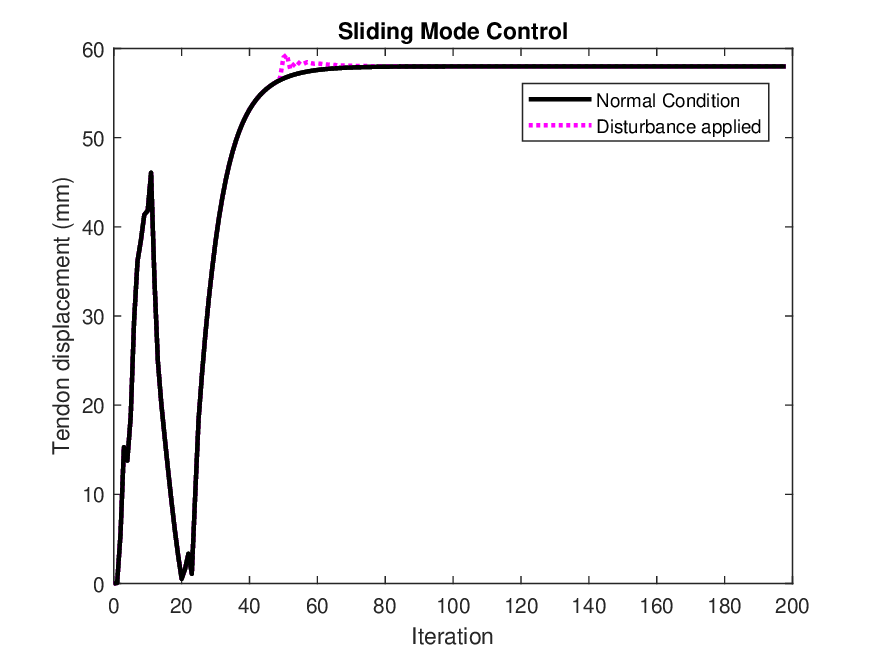}
        \caption{}
    \end{subfigure}
    \hfill
    \begin{subfigure}[b]{0.49\textwidth}
        \centering
        \includegraphics[width=\textwidth, trim={15 0 30 10}, clip]{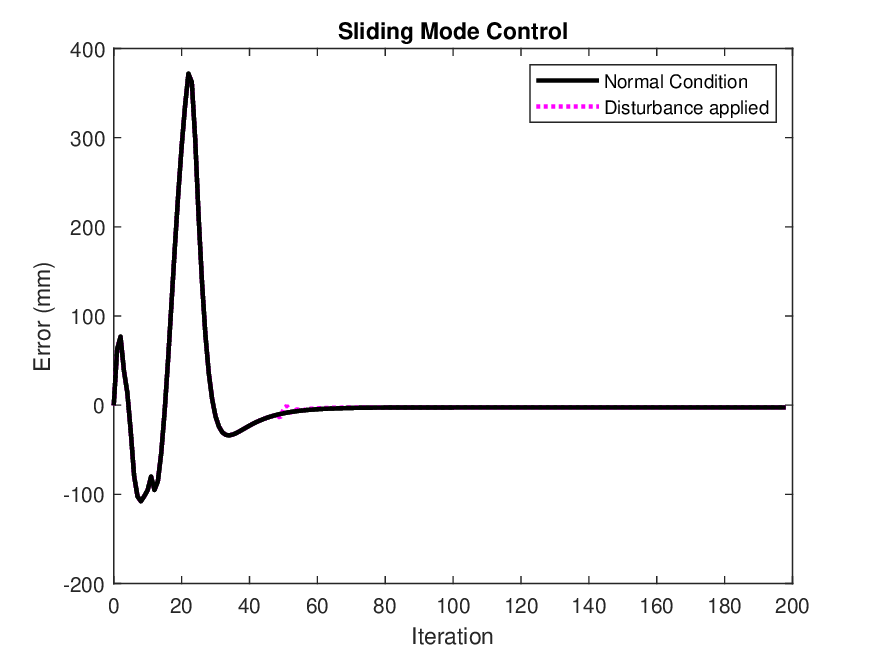}
        \caption{}
    \end{subfigure}
    \caption{Sliding mode control performance under disturbance scenario: (a) x-component of the robot's tip vs. iteration, (b) z-component of the robot's tip vs. iteration, (c) tendon displacement, and (d) position error.
}
    \label{Simulation_SMC_Disturbance}
\end{figure}

Figs. \ref{Simulation_BACK_Disturbance} and \ref{Simulation_SMC_Disturbance} illustrate the results of this scenario. A closer examination reveals that both controllers respond similarly to the applied disturbance, showing approximately equivalent performance.

To validate the simulated results, the experimental setup was employed. The position of the robot's tip relative to its base was recorded by a Vicon tracking system and imported to the Matlab codes as input. The controllers then calculate the error between the tip's position and the desired position, and determine the required tendon tension. Next, the tendon tension is converted to tendon displacement and provided to the motors to actuate the tendons. By rotating the motors, the position of the robot changes, and the CR repeats this loop until it reaches the desired position and the error is approached to zero. 

\begin{figure}[H]
    \centering
    \begin{subfigure}[b]{0.49\textwidth}
        \centering
        \includegraphics[width=\textwidth, trim={15 0 30 10}, clip]{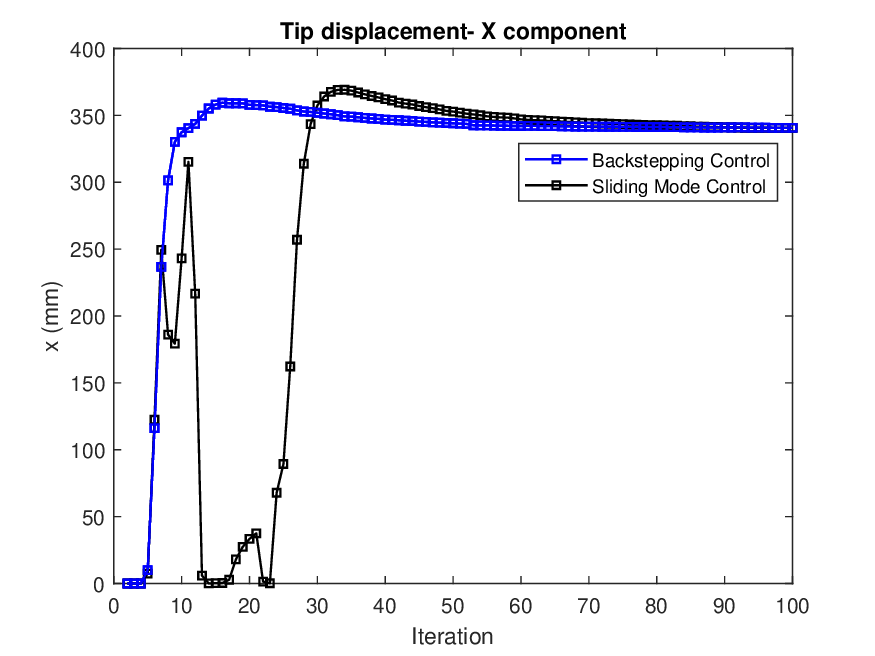}
        \caption{}
    \end{subfigure}
    \hfill
    \begin{subfigure}[b]{0.49\textwidth}
        \centering
        \includegraphics[width=\textwidth, trim={15 0 30 10}, clip]{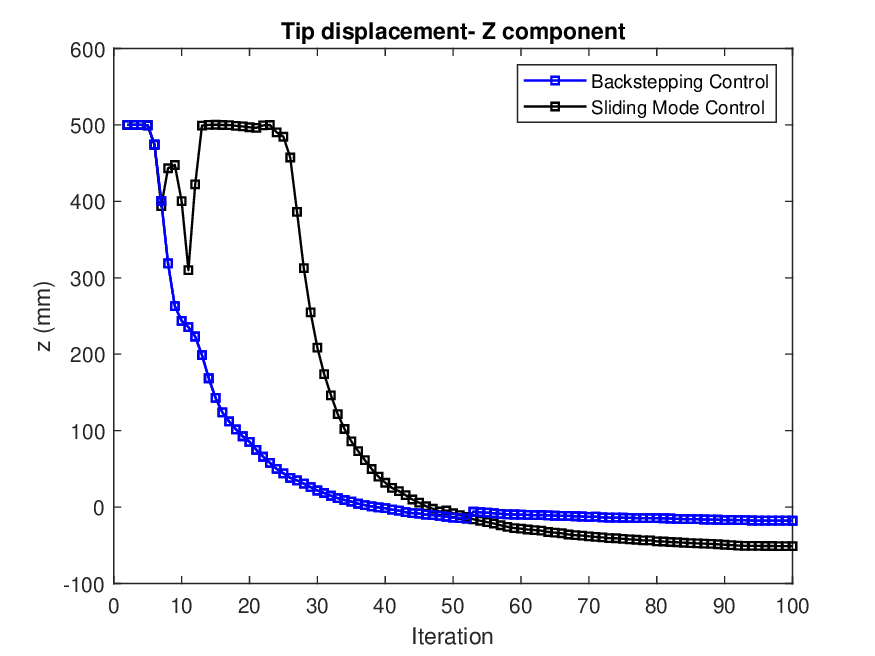}
        \caption{}
    \end{subfigure}
    \\
    \vspace{1em}
    \begin{subfigure}[b]{0.49\textwidth}
        \centering
        \includegraphics[width=\linewidth, trim={15 0 30 10}, clip]{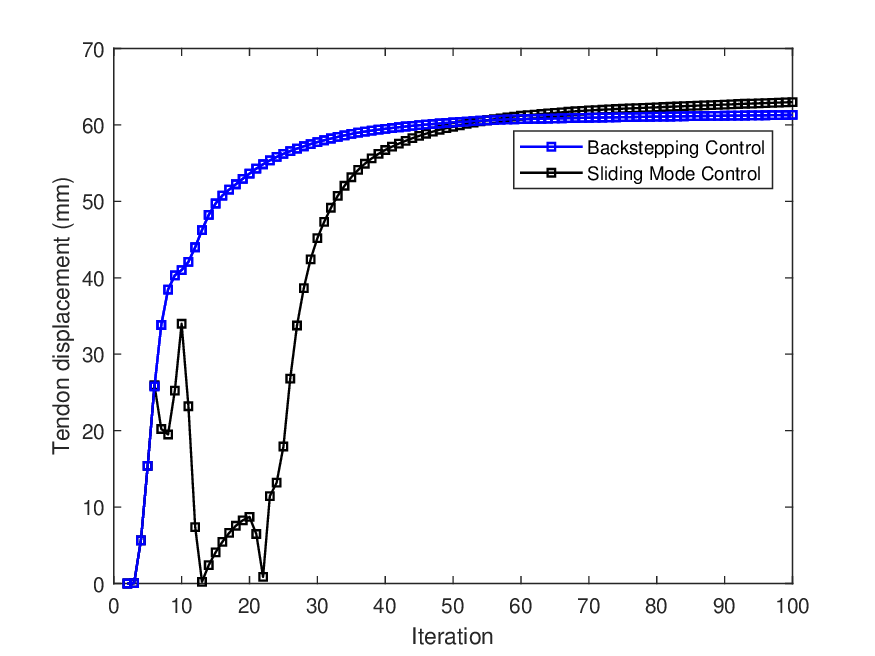}
        \caption{}
    \end{subfigure}
    \hfill
    \begin{subfigure}[b]{0.49\textwidth}
        \centering
        \includegraphics[width=\textwidth, trim={15 0 30 10}, clip]{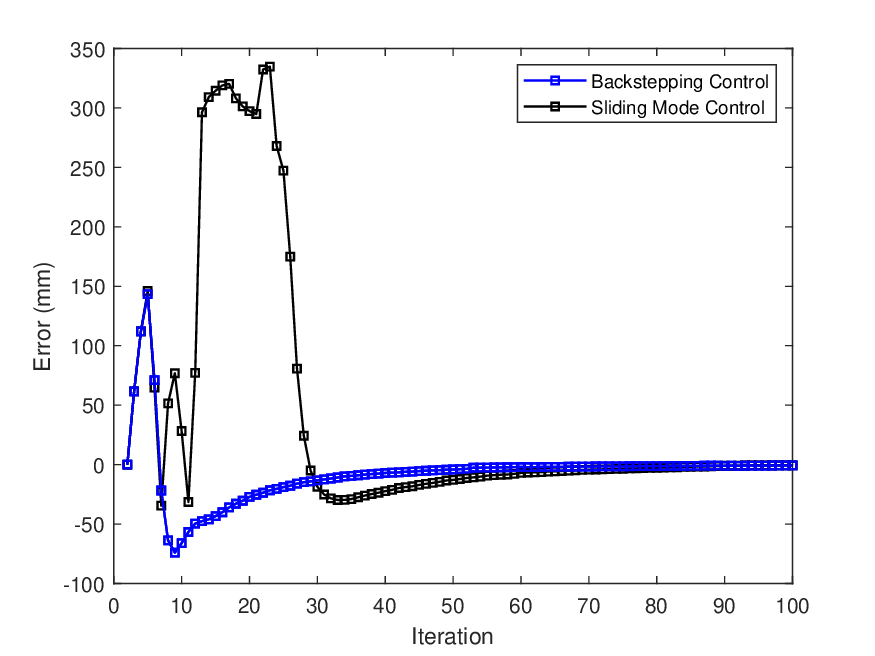}
        \caption{}
    \end{subfigure}
    \caption{Comparison of experimental results for backstepping and sliding mode control under normal conditions: (a) x-component of the robot's tip vs. iteration, (b) z-component of the robot's tip vs. iteration, (c) tendon displacement, and (d) position error.
}
    \label{Experiment_SMC_BACK_Normal}
\end{figure}

The experimental validation results shown in Figs. \ref{Experiment_SMC_BACK_Normal} (a) and (b) display the $x$ and $z$ components of the robot's tip. The backstepping control exhibits a smooth trajectory similar to the simulated results. In contrast, the sliding mode control showcases more fluctuations than its simulated counterpart.

Based on Fig. \ref{Experiment_SMC_BACK_Normal} (c), it can be concluded that the sliding mode control results in more significant changes in the tendon displacement. As a consequence, it may not be an ideal controller to employ in experiments, since it can lead to damage to the motor and structure of the CR.

Fig. \ref{Experiment_SMC_BACK_Normal} (d) illustrates the experimental error observed in both backstepping and sliding mode control methods. The error is a crucial indicator of the controller's performance, demonstrating how closely the robot's movements align with the desired trajectory. By analyzing the error signal, it can be seen that both controllers error approach zero, but with different behaviors.

The experimental quantitative comparison between backstepping control and sliding mode control under normal conditions is presented in Table \ref{Experiment}.
 In terms of smoothness, backstepping control demonstrates a significantly lower TPL of 379.07 mm, compared to 1243.16 mm for sliding mode control, indicating smoother movement. The settling time for backstepping control is notably shorter, with stabilization achieved within 21 iterations, while sliding mode control requires 44 iterations. Overshoot also favors backstepping control, with a lower percentage of 5.7\% versus 8.6\% for sliding mode control. Both controllers achieve a rise time from 0-90\%, with backstepping control responding in 7 iterations and sliding mode control in 10 iterations. Both methods attain zero steady-state error, demonstrating their ability to maintain accuracy once the system stabilizes. In conclusion, the experimental results demonstrate that backstepping control outperforms sliding mode control in terms of smoothness and settling time, as evidenced by the lower TPL and shorter settling time. Although both controllers achieve zero steady-state error, backstepping control exhibits lower overshoot and comparable rise time, making it a more effective choice.

\begin{table}[h!]
\centering
\caption{Experimental comparison of backstepping and sliding mode control under normal conditions.}
\label{Experiment}
\begin{tabular}{lcc}
\hline
\textbf{Criteria} & \textbf{Backstepping Control} & \textbf{Sliding Mode Control}\\
\hline
Smoothness & TPL= 379.07 mm & TPL= 1243.16 mm \\
Settling Time (5\%) & 21 iterations & 44 iterations \\
Overshoot & 5.7 \%  & 8.6 \% \\
Rise Time (0-90\%) & 7 iterations & 10 iterations \\
Steady-State Error & 0 & 0 \\
\hline
\end{tabular}
\end{table}

To further assess the controllers' performance in real-world conditions, we replicated two scenarios from the simulations in the experimental setup: one involving external weights and the other introducing disturbance forces.

\begin{figure}[H]
    \centering
    \begin{subfigure}[b]{0.49\textwidth}
        \centering
        \includegraphics[width=\textwidth, trim={15 0 30 10}, clip]{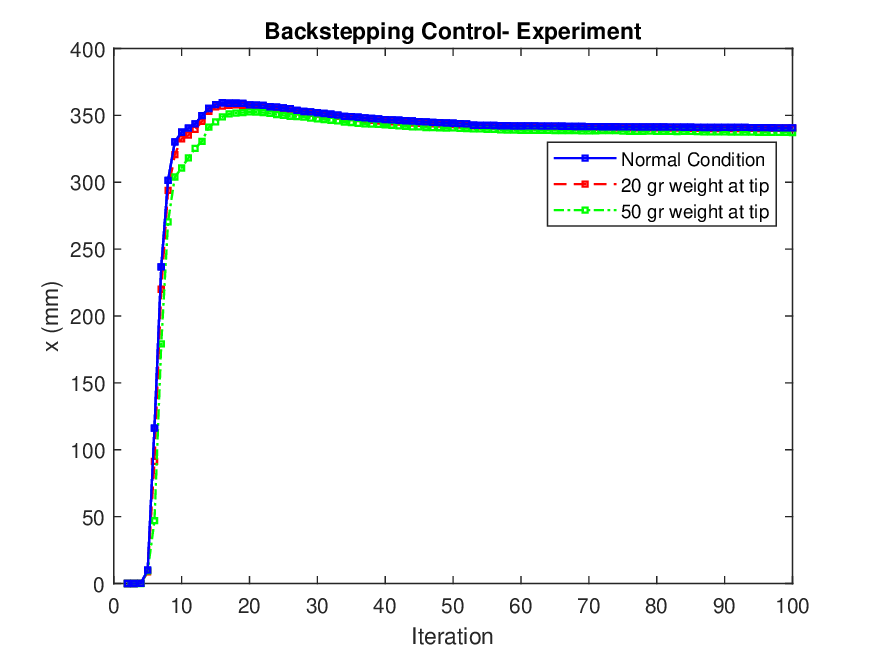}
        \caption{}
    \end{subfigure}
    \hfill
    \begin{subfigure}[b]{0.49\textwidth}
        \centering
        \includegraphics[width=\textwidth, trim={15 0 30 10}, clip]{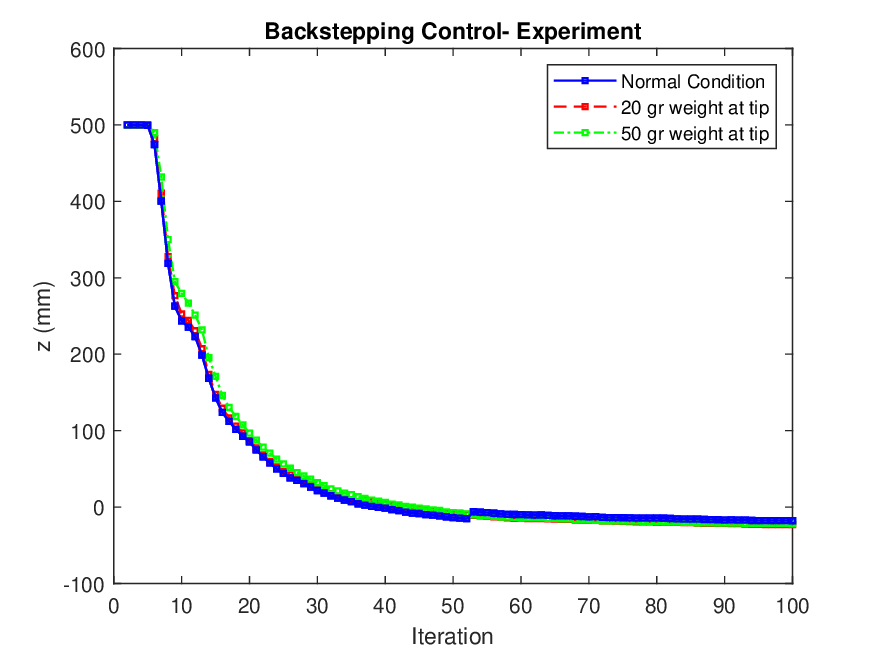}
        \caption{}
    \end{subfigure}
    \\
    \vspace{1em}
    \begin{subfigure}[b]{0.49\textwidth}
        \centering
        \includegraphics[width=\linewidth, trim={15 0 30 10}, clip]{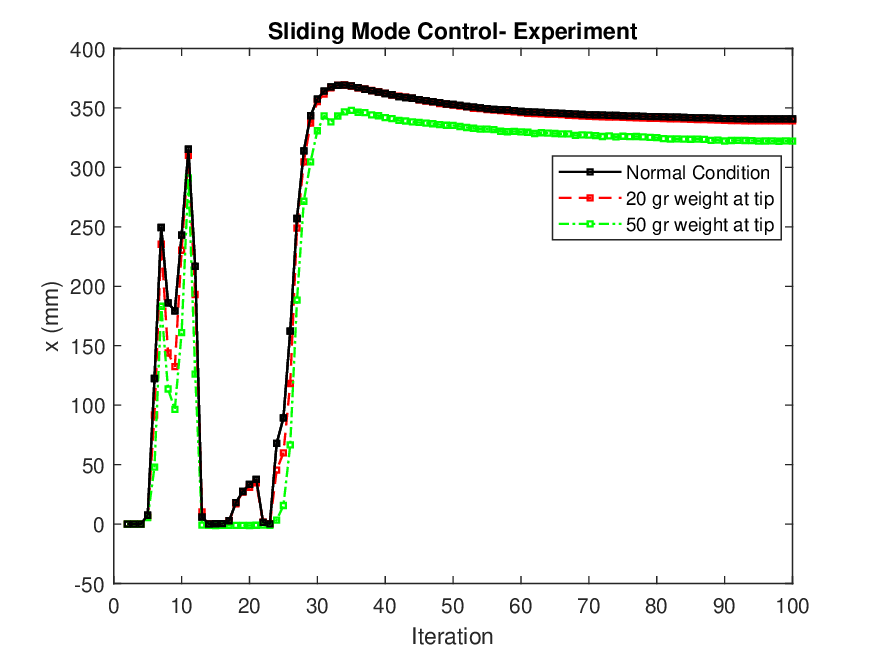}
        \caption{}
    \end{subfigure}
    \hfill
    \begin{subfigure}[b]{0.49\textwidth}
        \centering
        \includegraphics[width=\textwidth, trim={15 0 30 10}, clip]{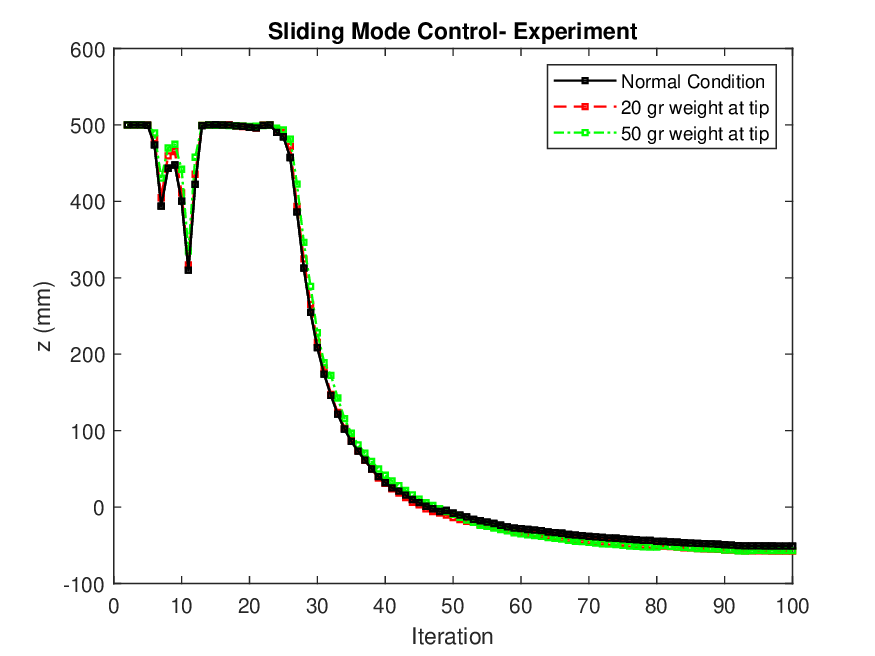}
        \caption{}
    \end{subfigure}
    \caption{Experimental comparison of backstepping and sliding mode controllers with hanging weight forces: (a) x-component in backstepping, (b) z-component in backstepping, (c) x-component in sliding mode, and (d) z-component in sliding mode.}

\label{Experiment_Both_weights}
\end{figure}

In the first experimental scenario, external forces were introduced by hanging weights of 20 g and 50 g at the tip of the robot. Unlike in the simulation, where the direction of the external force was kept constant, in the experimental setup, the direction of the force varied during the robot’s motion. In Fig. \ref{Experiment_Both_weights}, experimental results show the performance of backstepping and sliding mode controllers under external forces applied via hanging weights. Similar to the simulation results, sliding mode control exhibits more variation in response compared to backstepping control, indicating higher sensitivity to the external force.

\begin{figure}[H]
    \centering
    \begin{subfigure}[b]{0.49\textwidth}
        \centering
        \includegraphics[width=\textwidth, trim={15 0 30 10}, clip]{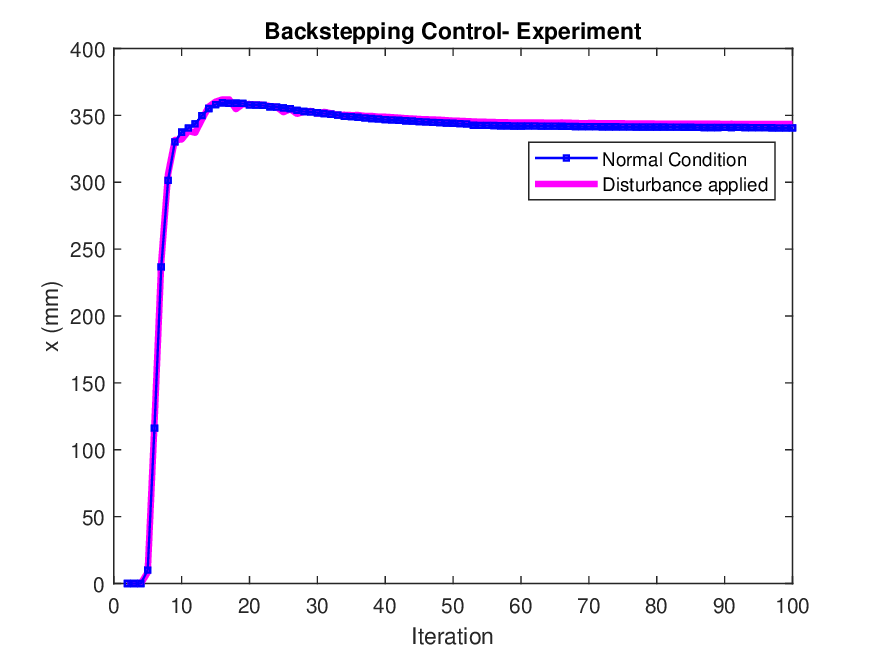}
        \caption{}
    \end{subfigure}
    \hfill
    \begin{subfigure}[b]{0.49\textwidth}
        \centering
        \includegraphics[width=\textwidth, trim={15 0 30 10}, clip]{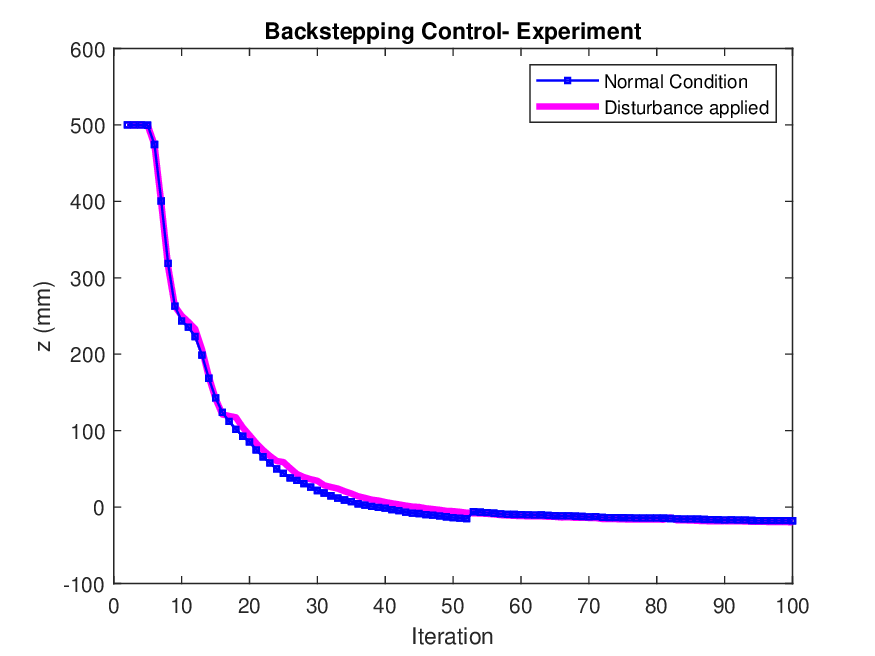}
        \caption{}
    \end{subfigure}
    \\
    \vspace{1em}
    \begin{subfigure}[b]{0.49\textwidth}
        \centering
        \includegraphics[width=\linewidth, trim={15 0 30 10}, clip]{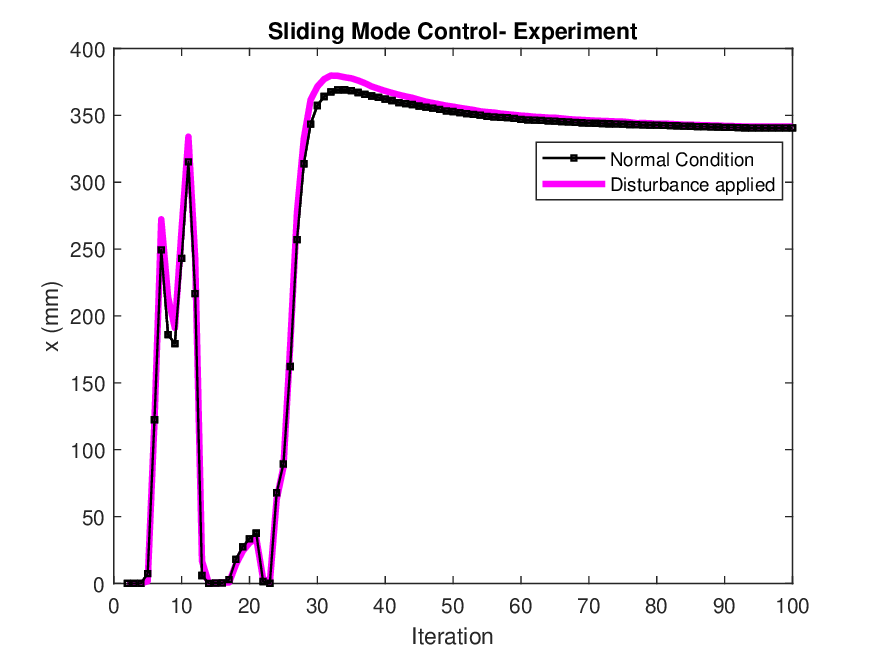}
        \caption{}
    \end{subfigure}
    \hfill
    \begin{subfigure}[b]{0.49\textwidth}
        \centering
        \includegraphics[width=\textwidth, trim={15 0 30 10}, clip]{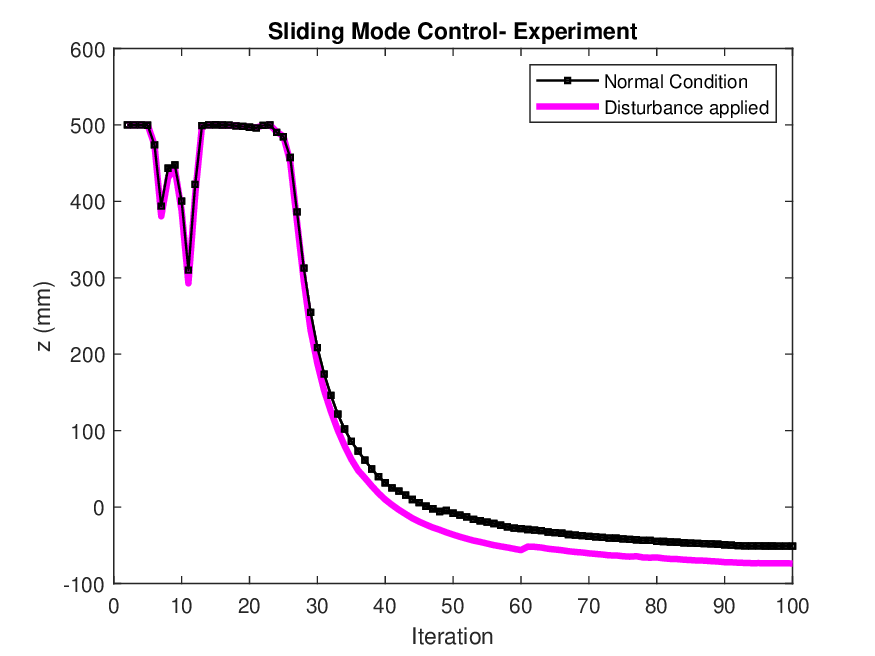}
        \caption{}
    \end{subfigure}
    \caption{Experimental comparison of backstepping and sliding mode controllers with applied disturbances: (a) x-component in backstepping, (b) z-component in backstepping, (c) x-component in sliding mode, and (d) z-component in sliding mode.}

\label{Expeiment_BOTH_disturbance}
\end{figure}

In the second experimental scenario, random disturbances were introduced to assess the controllers' performance under sudden, unexpected impacts. Disturbances with varying magnitudes and directions were applied at random locations along the robot’s structure at irregular intervals. Fig. \ref{Expeiment_BOTH_disturbance} demonstrates the performance of the controllers under applied impacts, revealing that sliding mode control experiences greater deviation compared to backstepping control.

\begin{figure}[H]
  \centering
  \includegraphics[width=0.95\textwidth]{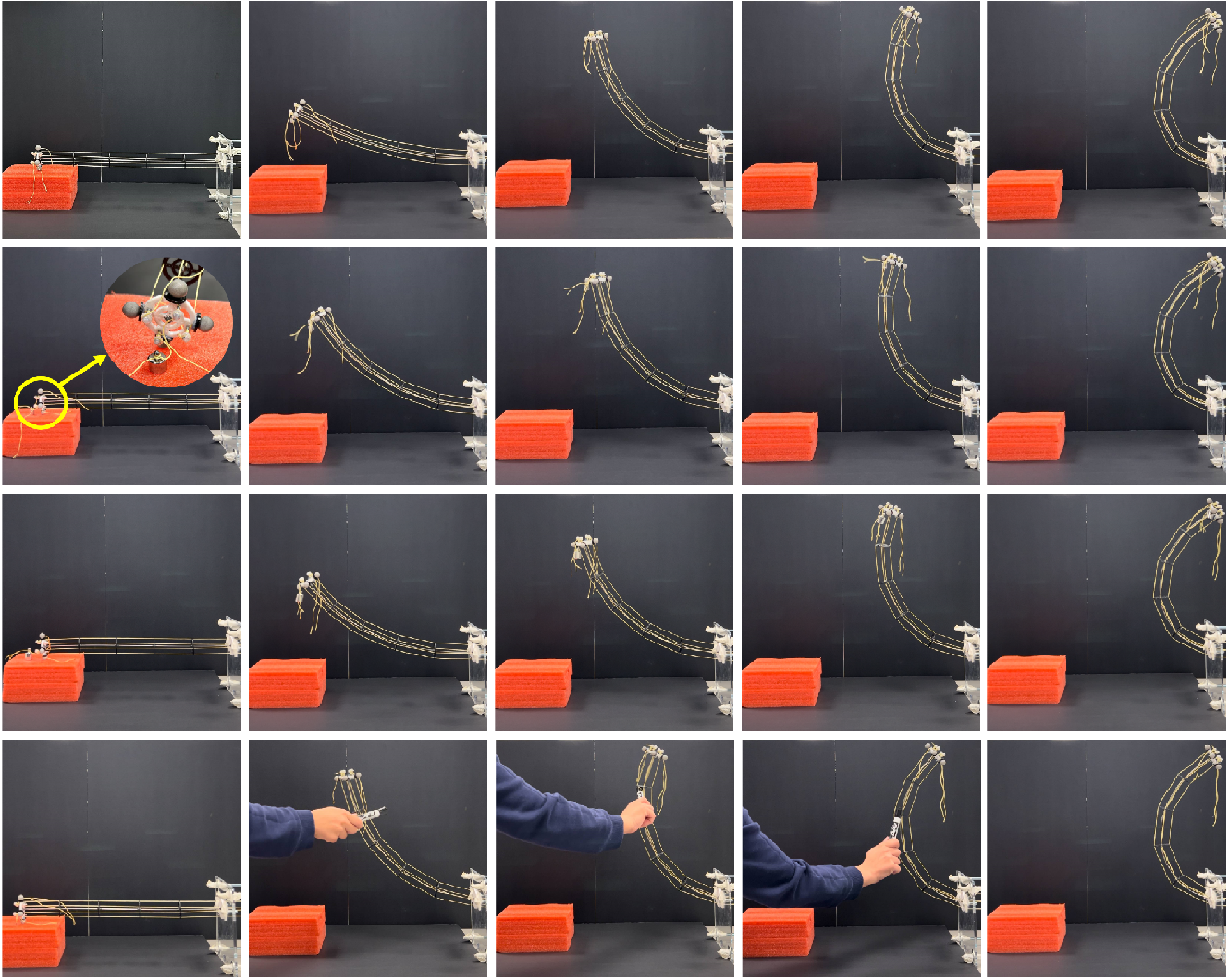}
  \caption{Snapshots of the TDCR under backstepping control: First row shows normal conditions, second row with a 20 g weight hanging from the tip, third row with a 50 g weight, and fourth row under random impact disturbances}
  \label{SnapShots_Back}
\end{figure}

\begin{figure}[H]
  \centering
  \includegraphics[width=0.95\textwidth]{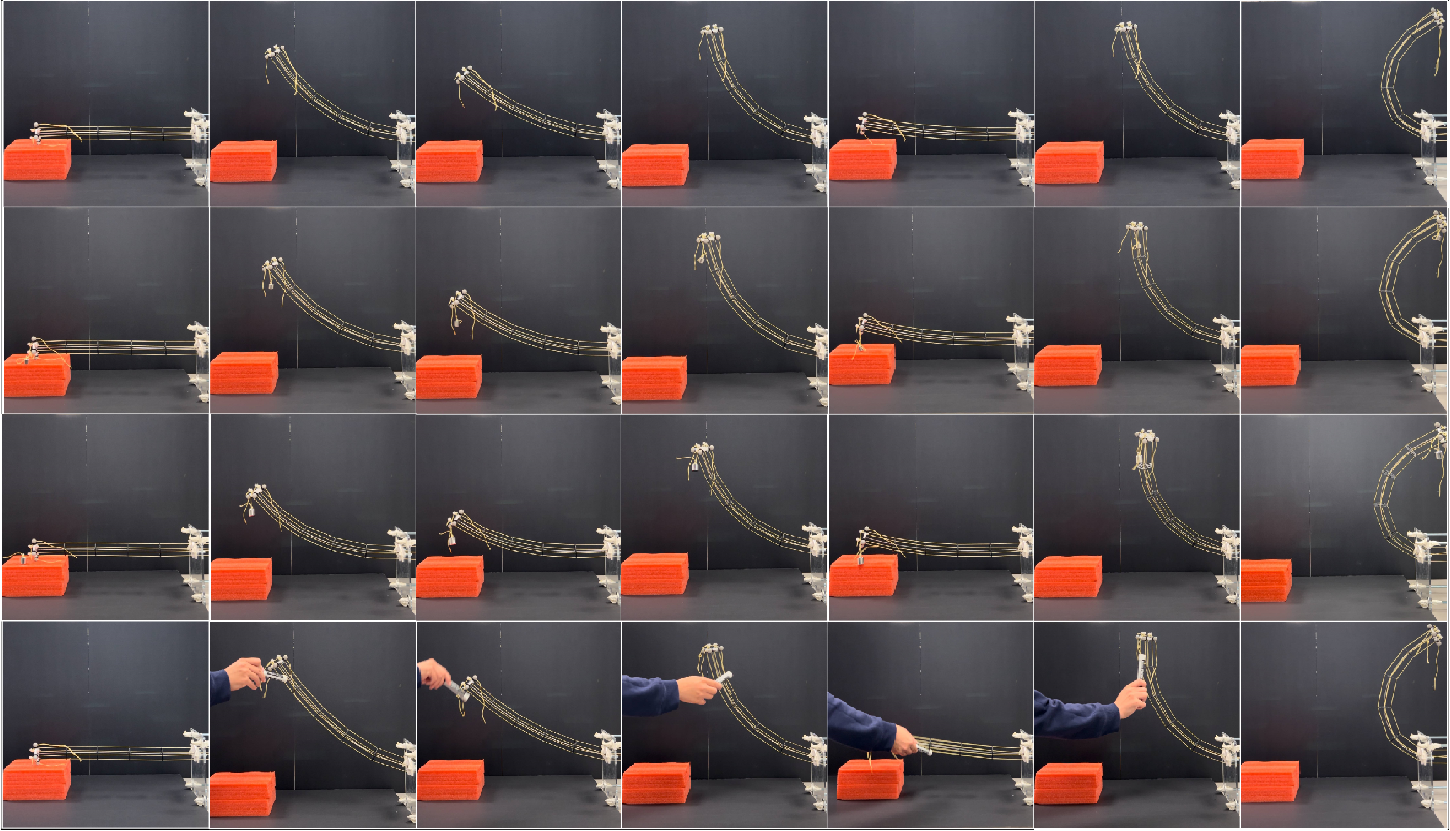}
  \caption{Snapshots of the TDCR under sliding mode control: First row shows normal conditions, second row with a 20 g weight hanging from the tip, third row with a 50 g weight, and fourth row under random impact disturbances}
  \label{SnapShots_SMC}
\end{figure}

In Figs. \ref{SnapShots_Back} and \ref{SnapShots_SMC}, the configuration of the TDCR under backstepping and sliding mode control is shown across various experimental scenarios. Each figure includes snapshots illustrating the robot's response under four distinct conditions.

The first row represents the normal conditions, showing the baseline performance of each controller.
The second and third rows display the TDCR’s behavior with a 20 g and 50 g weight hanged from its tip.
The fourth row illustrates the TDCR subjected to random impact disturbances, challenging the controllers to respond to unpredictable forces of varying magnitude and direction applied at different points along the robot.

\section{Conclusion}
\label{sec6}
In conclusion, the research presented in this paper successfully addressed the challenge of achieving large deflection in TDCRs through an experimentally validated approach. Unlike prior studies that focused on sliding mode control, this research proposes backstepping control as an alternative method to achieve large deflections with smoother trajectories, shorter settling time, reduced overshoot, and minimized deviation in the presence of external forces and disturbances.
 The simulation results demonstrated the capability of the control method in achieving large deflections while ensuring trajectory tracking and stability. These findings were further confirmed through experiments conducted on a physical prototype of the TDCR. Future research could explore the application of backstepping control in more complex CRs, such as parallel and co-manipulative ones.

\section{Acknowledgments}
This work was supported by the Natural Sciences and Engineering Research Council of Canada under Discovery Grant 2017-06930. Our sincere thanks go to Dr. Farhad Aghili for his valuable feedback and insightful comments, which significantly enhanced the quality of the manuscript.


\end{document}